\documentclass[journal]{IEEEtran}

\ifCLASSINFOpdf
\else
\fi

\usepackage[numbers]{natbib}
\usepackage{graphicx}
\usepackage{amsmath}
\usepackage{array}
\usepackage{caption}
\usepackage{booktabs}
\usepackage{tabularx}
\usepackage{multirow}
\usepackage{titlesec} 
\usepackage{hyperref}
\usepackage[font=small,labelfont=small,labelsep=period]{caption}
\usepackage{cite}
\usepackage{placeins}
\usepackage{orcidlink}
\usepackage{float}
\usepackage{gensymb}
\usepackage{enumitem}
\usepackage{makecell}

\newcommand{\tablefont}{\fontsize{8}{9}\selectfont}
\newcommand{\tablenotefont}{\fontsize{8}{9}\selectfont}

\newcolumntype{L}[1]{>{\raggedright\arraybackslash}p{#1}}
\newcolumntype{C}[1]{>{\centering\arraybackslash}p{#1}}
\newcolumntype{R}[1]{>{\raggedleft\arraybackslash}p{#1}}
\newcolumntype{Y}{>{\centering\arraybackslash}X}

\begin{document}

\title{Evaluating Semantic and Spatial Guidance for Foundation Model Segmentation of Small-Scale PV in Remote Sensing Imagery}

\author{
    Roni~Blushtein-Livnon\raisebox{0.5ex}{\orcidlink{0000-0002-3493-4894}}, 
    Tal~Svoray\raisebox{0.5ex}{\orcidlink{0000-0003-2243-8532}}, Osher~Rafaeli\raisebox{0.5ex}{\orcidlink{0000-0002-7097-7568}}, Michael~Dorman\raisebox{0.5ex}{\orcidlink{0000-0001-6450-8047}}, Itay~Fischhendler\raisebox {0.5ex}{\orcidlink{0000-0003-2243-8532}}, Havazelet~Yahel\raisebox {0.5ex}{\orcidlink{0000-0003-2243-8532}}, Emir~Galilee\raisebox{0.5ex}{\orcidlink{0000-0001-7892-4172}}

    \thanks{R. Blushtein-Livnon and M. Dorman and O. Rafaeli are with the Department of Environmental, Geoinformatics and Urban Planning Sciences, Ben-Gurion University of the Negev, Israel (e-mail: livnon@bgu.ac.il; dorman@bgu.ac.il).}
    \thanks{T. Svoray is with the Department of Environmental, Geoinformatics and Urban Planning Sciences, and the Department of Psychology, Ben-Gurion University of the Negev, Israel (e-mail: tsvoray@bgu.ac.il).}%
    \thanks{I. Fischhendler is with the Department of Geography, The Hebrew University of Jerusalem, Isreal}%
    \thanks{H. Yahel and E. Galilee are with the Ben Gurion Institute for the Study of Israel \& Zionism, Ben-Gurion Israel Research Institute, Isreal}%
   
}

\maketitle
\begin{abstract}
Spatio-temporal PV data are essential for understanding adoption processes in off-grid regions, yet such data remain largely unavailable. Automated segmentation of remote sensing (RS) imagery offers a promising solution; yet, residential PV systems remain challenging targets because of their small size and sparse distribution, resulting in severe target-background imbalance. Vision-language foundation models (FMs) provide a data-efficient paradigm through prompt-based semantic and spatial guidance, but the relative contribution of different prompt types remains unclear. We systematically evaluate SAM3 for small-scale PV segmentation in RS imagery by comparing textual, geometric, and hybrid prompting, under varying supervision levels, training strategies, spatial resolutions, and imaging conditions. Multi-temporal aerial imagery from a large off-grid rural region serves as a study site, with findings validated across three additional datasets. Prompting strategy emerged as the dominant factor governing model behavior. Textual prompting consistently produced the lowest performance and showed the greatest sensitivity to supervision and imaging conditions. In contrast, spatial guidance substantially improved both segmentation accuracy and robustness. Hybrid prompting achieved the highest accuracy and stability, indicating that semantic and spatial guidance provide complementary information. Most performance gains were achieved with only a few hundred annotated samples, demonstrating strong data efficiency. Transfer learning had limited overall impact, with only modest improvements observed for textual prompting under limited supervision. Overall, our findings establish prompting strategy as a key determinant of SAM3 adaptation, robustness, and generalization, highlighting the potential of promptable FMs for scalable PV mapping in data-constrained off-grid regions.
\end{abstract}


\IEEEpeerreviewmaketitle

\section{Introduction}
Despite rapid growth in solar energy deployment, $\sim$730 million people world wide remained without electricity in 2024 \citep{IEA2025_population}, making the acceleration of PV adoption in off-grid regions a pressing social priority. Achieving this goal requires detailed knowledge of the location, extent, and evolution processes of PV adoption \citep{Graziano2015, Mahn2024, de2016heterogeneity, BaltaOzkan2021}. Yet, such information remains scarce. Monitoring efforts in these regions typically rely on incomplete market records or small surveys rather than comprehensive inventories \citep{IRENA_2025, GOGLA2023}, constraining both research and policy development precisely where understanding and promoting PV diffusion matters most for welfare and socioeconomic development.

Recent advances in deep learning enable automated segmentation of PV installations from RS imagery, facilitating construction of fine-grained spatio-temporal datasets. Yet, the task is especially challenging in off-grid settings, where PV systems are small and spatially dispersed, resulting in severe target-background imbalance conditions \citep{guo2024transpv}.

Automated segmentation of residential PVs has predominantly relied on specialized architectures that are often optimized for specific datasets \citep{wang2025pv, guo2024transpv, tan2023enhancing}. While highly effective, these approaches typically require large annotated training data and may generalize poorly in data-constrained conditions or out-of-distribution settings \citep{tan2024general}.

The emergence of foundation models (FMs) has introduced a new paradigm for image segmentation \citep{huo2025remote}. By leveraging large-scale pretraining and flexible prompt-driven interaction, FMs may offer improved generalization and reduced dependence on extensive task-specific annotation \citep{kirillov2023, chen2024rsprompter, chen2025multiscale}. Yet, adapting FMs to RS imagery is still under intensive research, as most FMs are pretrained on natural images and face a substantial domain gap \citep{yao2025remotesam, zhang2025rsam, wang2023samrs}. 

Recent advances in vision-language models (VLMs) have expanded how FMs can be guided, enabling segmentation through natural-language descriptions rather than explicit spatial inputs \citep{radford2021learning, alayrac2022flamingo}. Unlike geometric prompts, which often require manual specification or auxiliary models \citep{jiang2026}, textual prompting offers a more scalable path to automated segmentation with minimal supervision \citep{zhang2024segclip, li2025segearth, huo2025remote}. Embedding vision-language capabilities in FMs raises the possibility of performing segmentation using different forms of guidance, ranging from explicit spatial cues to high-level semantic descriptions. In the RS domain, recent RS-VLMs have demonstrated effective alignment between imagery and natural language, enabling semantic retrieval, cross-modal understanding, and open-vocabulary scene interpretation \citep{liu2024remoteclip, zhang2024rs5m}. These advances suggest that language can serve as an effective source of guidance for downstream vision tasks and have stimulated growing interest in extending language-driven interaction beyond image-level analysis.

Despite these promising developments, several important research gaps remain. Whether language guidance derived from image-level RS understanding can effectively support precise dense prediction tasks, rather than coarse scene interpretation, is still an open question \citep{li2025segearth}. The challenge is fundamental: unlike geometric prompts, which provide explicit spatial constraints that directly anchor segmentation, semantic descriptions require the model to infer object localization from semantic cues distributed across the entire image, without any spatial reference. This distinction raises important questions about the relative contribution of each guidance modality and how well either performs under heterogeneous imaging conditions, resolutions, and acquisition settings, that characterize real-world RS data \citep{jiang2026}. These uncertainties become particularly consequential in settings defined by small objects and severe target-background imbalance, where reliable segmentation hinges on accurate localization despite limited and often ambiguous visual evidence \citep{yao2026pvsam, li2025segearth}. 

More specifically, the factors governing the adaptation, robustness, and generalization of promptable FMs, such as SAM3, under such challenging conditions, remain poorly understood, particularly for small-object segmentation under RS domain shift. Moreover, the individual effects of different prompting modalities are not systematically disentangled, leaving a critical gap between the promise of FMs for RS analysis and their demonstrated reliability in operationally relevant scenarios.

To address these gaps, we systematically evaluate the effectiveness of SAM3, a SOTA FM, for small-scale PVs segmentation in RS imagery. An emphasis is placed on understanding the relative roles of semantic and spatial guidance, and on assessing how prompting strategies, supervision scale, training strategies, and imaging conditions jointly influence segmentation performances. The evaluation is conducted primarily in a vast off-grid rural setting using multi-temporal aerial imagery, and is subsequently extended to assess the consistency of observed patterns across diverse datasets and environmental contexts. Accordingly, the study addresses the following research questions:

\begin{enumerate}[label=\arabic*., leftmargin=*, labelsep=0.5em, itemsep=-0.52pt, topsep=2pt]
\item \textit{Prompting strategy:}
To what extent do different forms of guidance, including textual, geometric, and hybrid prompting, influence the performances of small-scale PV segmentation?
\item \textit{Training regime:}
How do supervision scale and training strategy, including independent training and incremental transfer learning, influence the performance of different prompting strategies?
\item \textit{Imaging conditions:}
To what extent are different prompting strategies influenced by variations in imaging conditions, including spatial resolution and cross-image variability?
\item \textit{Cross-dataset consistency:}
To what extent are the observed performance patterns consistent across datasets representing diverse environmental and contextual conditions?
\end{enumerate}

\section{Related work} \label{background}
\subsection{PV segmentation: challenges and existing approaches}
Accurate segmentation of small and dispersed objects in RS imagery is challenging due to severe target-background imbalance, a problem well exemplified by residential PV installations \citep{guo2024transpv, li2021understanding}. PV systems also exhibit substantial geometric variability \citep{jiang2021multi, li2021understanding, yao2026pvsam}, including differences in shape, orientation, inclination, and spatial configuration. They may range from compact to elongated forms, produce diverse shadow patterns due to varying tilt angles, and occur on both rooftops and ground surfaces, resulting in varying levels of contrast with the surrounding background \citep{garcia2024generalized}. Heterogeneity in these characteristics is particularly pronounced in less formal installation contexts, i.e., off-grid rural environments. Detection is further hindered by radiometric ambiguity, as PV panels often share spectral and textural properties with surrounding materials, such as rooftops, glass, and water \citep{li2025joint}, leading to frequent false positives and ambiguous boundaries \citep{li2024review}. Sensitivity to illumination and acquisition timing further introduces substantial variation in color and shadow patterns, undermining feature stability and limiting model generalization \citep{li2021understanding, tan2024general}.

These challenges prompted a transition from traditional machine learning methods (e.g., \citep{malof2016}) to deep learning-based frameworks. Initial CNN-based approaches often relied on weak supervision derived from image-level labels. For example, DeepSolar \citep{yu2018deepsolar} combines image classification with activation maps to infer panel locations, enabling large-scale mapping but providing only coarse localization rather than accurate delineation. Subsequent work adopted encoder-decoder CNN architectures for supervised semantic segmentation, such as FCN, U-Net, and SegNet, with models like DeepLabv3+ further improving performance \citep{kausika2021geoai, camilo2018application}. Building on these advances, frameworks such as SolarNet \citep{hou2019solarnet} achieved strong results for large-scale solar farms. However, since they were developed primarily for utility-scale installations, they are less suitable for small and spatially fragmented PV systems \citep{garcia2024generalized, adib2025deep}. Later studies introduced task-specific adaptations to enhance feature representation and robustness. These include attention modules, spatial-context modeling, multi-scale feature learning strategies, and the integration of transfer learning and pretrained backbones to improve feature representation and robustness under diverse imaging conditions \citep{guo2024transpv, tan2023enhancing, lu2024pv, zhao2025enhancing}.

A complementary line of work focuses on improving boundary delineation and fine structural representation. Boundary-aware approaches incorporate edge detection \citep{li2025joint}, shape constraints \citep{tan2023enhancing}, and refinement modules to improve contour accuracy and mitigate boundary ambiguity \citep{guo2024transpv, wang2025pv}. More recent studies extend this direction through joint-task learning, integrating segmentation and edge detection to improve localization \citep{li2025joint}. Additional advances include scale-adaptive and position-guided mechanisms designed to better capture PV structures with varying geometries \citep{li2025joint}. Complementary approaches leverage frequency-domain representations to improve sensitivity to high-frequency components such as edges and fine structural details \citep{wang2025pv}.

Recent transformer-based architectures such as SegFormer \citep{xie2021segformer} and Mask2Former \citep{cheng2022masked} have advanced semantic segmentation through improved modeling of long-range spatial dependencies and multi-scale representations. Regarding PVs, recent studies demonstrate the superior performance of these architectures over CNN-based models across multi-resolution and heterogeneous datasets \citep{garcia2024generalized}. For example, TransPV combined a U-Net-style encoder-decoder architecture with Vision Transformer components to jointly model global context and fine spatial details, improving robustness to the diverse structures of residential PV installations \citep{guo2024transpv}. 

Despite these important advances, such approaches remain dependent on task-specific architectures, large annotated datasets, and extensive model adaptation, limiting their scalability and generalizability across diverse environmental settings, imaging conditions, and low-data scenarios \citep{zech2024toward}. These limitations motivate the exploration of more data-efficient and generalizable modeling paradigms for PV segmentation.
\vspace{-8pt}
\subsection{FMs for RS segmentation}
Recent advances in CV-driven RS analysis have increasingly explored the adoption of FMs \citep{huo2025remote}, which offer strong generalization capabilities across a wide range of tasks \citep{chen2024rsprompter}. Notably, SAM and alike have demonstrated impressive ZS segmentation performance, enabling segmentation without task-specific training \citep{kirillov2023}. However, as FMs are predominantly trained on natural image datasets, their application to RS imagery introduces a pronounced domain gap, limiting their ability to generalize under out-of-distribution conditions \citep{chen2025multiscale, yao2025remotesam}. This limitation stems from fundamental differences between RS imagery and natural images, including object density, complex spatial organization, high intra-class variability, and substantial scale variation \citep{zhang2025rsam, yao2025remotesam}. Achieving robust segmentation in RS imagery is further exacerbated by frequent presence of low-density, small, and low-contrast targets \citep{wang2023samrs, ren2024segment, chen2024rsprompter}. Task-specific adaptation and fine-tuning (FT) are therefore often necessary for effective FM deployment in RS \citep{liu2025pointsam, zhu2017deep}, raising questions about whether their promised data efficiency can actually be realized in data-constrained settings.

To address these domain-specific challenges, recent studies have proposed three adaptation strategies for applying FMs to RS segmentation. A \textit{first} line of work focuses on domain-specific FT and architectural adaptations to improve the representation of multiscale spatial patterns. These include integrating scale-adaptive modules and edge-aware components to enhance delineation of complex, heterogeneous structures \citep{chen2025multiscale, zhang2025rsam, chen2025edge}. A \textit{second} line of research emphasizes data-efficient model adaptation, seeking to reduce reliance on large annotated datasets. These include weakly supervised and self-training frameworks that leverage sparse annotations, such as points or pseudo-labels, to guide model adaptation \citep{liu2025pointsam}. A \textit{third} and rapidly growing direction seeks to exploit the prompt-driven interaction capabilities of FMs, particularly SAM, through spatial guidance such as points, bounding boxes, and masks. These approaches utilize such prompts to improve localization performance, often by automatically generating task-relevant prompts or deriving them from auxiliary models such as object detectors, thereby reducing manual intervention \citep{chen2024rsprompter, osco2023segment, ren2024segment, diab2025optimizing}. 
However, prompt-based methods remain highly sensitive to prompt quality and frequently depending on auxiliary models, or manual inputs, for geometric prompt generation, limiting their robustness and scalability as fully automated segmentation pipelines.

Yet, the literature lacks a systematic evaluation of training data requirements, as most studies prioritize segmentation performance over explicitly quantifying the role of supervision \citep{liu2025pointsam, chen2024rsprompter}. While SAM-based adaptation achieved strong performance under limited supervision in domains such as medical imaging \citep{piater2025prompt, cheng2024hsam}, comparable evidence remains scarce in RS. Consequently, the extent to which FM performance depends on supervision scale remains poorly understood, particularly for small-object segmentation under severe class imbalance.
\vspace{-8pt}
\subsection{VLMs and prompt-based segmentation}
Existing FM approaches rely primarily on visual information and, even when enhanced by attention mechanisms and geometric prompting, they remain largely anchored to visual similarity and spatial context, without explicit access to higher-level semantic knowledge \citep{zhang2024segclip}. Reliance on visual information is restrictive in RS imagery, where high inter-class visual similarity and context-dependent interpretation hamper object recognition, and visual appearance alone may be insufficient to distinguish between semantically different objects \citep{jiang2026}. This has motivated growing interest in leveraging language-based representations to enrich visual features with semantic context and improve scene understanding \citep{zhang2024segclip}. Among these developments, VLMs have emerged as a framework that learns joint image-language representations and integrates semantic knowledge into visual recognition tasks.

Extending this paradigm to RS, RS-VLMs models, such as GeoRSCLIP and RemoteCLIP, demonstrate that meaningful semantic alignment between language and RS imagery is feasible, enabling ZS classification, cross-modal retrieval, and semantic localization \citep{li2024vision, zhang2024rs5m, liu2024remoteclip}. More recent architectures further extend these capabilities through richer textual descriptions and improved modeling of spatial-semantic relationships \citep{chen2025dgtrsd}. These advances provide a semantic foundation for extending language-guided interaction beyond image-level understanding, leading to open-vocabulary semantic segmentation (OVSS), which leverages VLM-encoded semantic knowledge to support segmentation beyond predefined categories while reducing reliance on exhaustively labeled datasets \citep{li2025segearth, xin2025segearth}. Building upon OVSS, recent developments have extended VLM-based segmentation toward instruction-driven frameworks, in which natural language descriptions provide richer semantic and contextual guidance for pixel-level prediction \citep{jiang2026, xin2025segearth}. More advanced approaches further incorporate language-guided reasoning, enabling models to address increasingly complex semantic and spatial queries \citep{xin2025segearth}. In parallel, the construction of large-scale image-text datasets tailored to RS has further improved semantic alignment and domain adaptation, supporting more robust cross-domain generalization \citep{yuan2025}.

However, despite their strong semantic capabilities, VLM-based approaches continue to struggle with dense prediction tasks \citep{zhou2024geo}. Most VLMs are optimized for image-level alignment rather than pixel-level prediction, and their representations consequently lack the spatial granularity required for precise object segmentation in RS imagery \citep{li2025segearth, zhou2024geo}, manifesting as coarse boundaries and degraded localization accuracy \citep{li2025segearth}. This reflects a fundamental tension between semantic and spatial forms of guidance. Unlike image retrieval or scene understanding, segmentation demands accurate localization and boundary delineation at the pixel level. In text-driven settings, models must infer object location from textual descriptions alone, effectively conducting a global semantic search across the image with no explicit spatial reference. Geometric prompting, by contrast, provides direct spatial constraints that confine the search to predefined regions of interest. These structural differences suggest that semantic and spatial guidance may respond quite differently to variations in target characteristics, imaging conditions, and scene context. Such differential sensitivity has received little attention to date.

To compensate for these limitations, existing approaches incorporate additional spatial cues, auxiliary modules, structural priors, or external refinement models to improve localization accuracy \citep{jiang2026, li2025segearth, sultan2023geosam}. While often effective, such solutions increase pipeline complexity and may introduce additional supervision requirements or dependencies on auxiliary models. Examples include GeoGround \citep{zhou2024geo} and SegEarth-OV \citep{li2025segearth}, which integrate semantic representations from VLMs with spatial refinement mechanisms to support localization or segmentation tasks. However, the extent to which semantic and spatial cues can effectively substitute or complement one another remains an open question, and direct comparisons between semantic, spatial, and combined guidance approaches are limited.

Recent FMs, such as SAM3, introduce a promptable segmentation paradigm that bridges large-scale visual pretraining with semantic conditioning. Rather than relying solely on image-text alignment, SAM3 uses short noun phrases as semantic concepts that directly condition both recognition and pixel-level segmentation, enabling delineation of multiple instances associated with a given concept within a scene \citep{carion2025sam3}. This capability is supported by a query-driven design that jointly models concept presence and mask generation. While this formulation suggests strong potential for ZS generalization and reduced reliance on dense annotations, its effectiveness under domain shift has yet to be established. 

Although recent studies have demonstrated the potential of SAM3 and related vision-language foundation models for remote-sensing segmentation, they have primarily focused on engineering adaptations tailored to specific prompting schemes \citep{li2025segearth}. Existing research remains fragmented across separate text-guided and geometry-guided paradigms, with little understanding of how alternative forms of guidance influence model behavior and no comparison between them \citep{zhang2024T2S, yao2026pvsam}. Consequently, the role of prompting modality in shaping FM performance under RS domain shift remains largely unexplored.

\section{Methods and Materials}
\subsection{Study area and primary dataset}
Multi-temporal RGB aerial imagery of the Northern Negev, Israel \citep{bluestein2023economic}, was acquired across seven years (Table \ref{res_year}). The imagery was obtained from the \textit{Authority for Development and Settlement of the Bedouin in the Negev} and the \textit{Survey of Israel}. The image series captures substantial variability in imaging conditions, including differences in spatial resolution and radiometric properties across years. It also inherently reflects temporal variation in scene content, including changes in the extent and spatial distribution of the built environment and PV installations, and seasonal variation in vegetation cover associated with acquisition timing.

\begin{table}[t]
\centering
\caption{\textbf{Target characteristics and dataset extent by spatial resolution.}}
\label{res_year}
\footnotesize
\renewcommand{\arraystretch}{1.15}
\setlength{\tabcolsep}{3pt}

\begin{tabular*}{\columnwidth}{@{\extracolsep{\fill}}cccccc}
\toprule
\makecell{\textbf{Spatial}\\\textbf{resolution}\\(m/pixel)} &
\makecell{\textbf{Median}\\\textbf{PV size}\\(pixel)} &
\makecell{\textbf{PV size}\\\textbf{IQR}\\(pixel)} &
\makecell{\textbf{Acquisition}\\\textbf{years}} &
\makecell{\textbf{Annotated}\\\textbf{PVs}\\(count)} &
\makecell{\textbf{Annual}\\\textbf{mapped}\\\textbf{area} (km\textsuperscript{2})} \\
\midrule

0.250 & 69  & 43-108  & 2012, 2015, 2016 & 8,032 & 5.73 \\
0.145 & 205 & 127-321 & 2017, 2020       & 5,540 & 1.93 \\
0.132 & 248 & 153-388 & 2021, 2022       & 6,791 & 1.60 \\

\bottomrule
\end{tabular*}

\begin{minipage}{\columnwidth}
\vspace{3pt}
\footnotesize
\textbf{Note:} Statistics are computed from the complete annotated dataset. Median and IQR values are derived from the pooled distribution of PV areas across all acquisition years and converted to pixels. Total PVs represent the cumulative number of annotated installations across all acquisition years within each resolution group. Mapped area refers to the ground area covered by the tiles available for each acquisition year.
\end{minipage}
\end{table}

The study area covers $\sim$1,150 km² and includes $>$1,100 off-grid settlement clusters of the Bedouin population, a traditional rural ethnic minority. The built environment is characterized mainly by lightweight metal structures. The clusters are spatially dispersed across the landscape and exhibit low internal residential density. PVs are installed either on rooftops or on the ground, with varying tilt angles and orientations and without standardized configurations. 

Table~\ref{res_year} summarizes both target characteristics and dataset extent across spatial resolutions, including median object size in pixels, the mapped area in each acquisition year, and the total number of annotated PVs. Across the annotated dataset, PV installations were generally small, with a median area of 4.33 m$^2$ (IQR = 2.67-6.76 m$^2$). The corresponding median PV size in pixels demonstrating how the level of spatial detail available for PV segmentation varies with spatial resolution. Across the entire dataset, PV pixels accounted for $\sim$0.48\% of all pixels, and corresponding to a foreground-background ratio of $\sim$1:207, indicating severe class imbalance.
\vspace{-8pt}
\subsection{Annotation protocol and data preprocessing}
Ground truth annotations were generated by three expert annotators following a structured protocol described in \citet{blushtein2025performance}. Each image was independently annotated by all three experts, and the final reference annotation was derived through pixel-level majority voting. Previous evaluation showed that this protocol produced higher-quality annotations than either sequential review-based annotation or individual annotation \citep{blushtein2025performance}. The resulting consensus masks served as the gold standard throughout the study.

All images were subsequently preprocessed to ensure consistent spatial representation and pixel value scaling. This included tiling the imagery into fixed-size patches (256×256 pixels), selected to match the spatial scale of the targets, along with normalization of pixel values and alignment where necessary. Each tile consisted of an image–mask pair, where the binary mask delineates PV installations at the pixel level. Tiles were generated as GeoTIFF files, preserving geospatial metadata, enabling the reconstruction of PV installations mapping from model predictions.

Object-level geometric representations for geometric and hybrid prompting were derived from annotated masks as minimal enclosing rectangles and stored in text files containing bounding box (BB) coordinates and object centroids.

The dataset was partitioned separately for each acquisition year using geographically non-overlapping regions to prevent spatial leakage. Each yearly dataset consisted of 200 validation samples, 500 test samples and 700 training samples. Validation and test partitions were held fixed throughout all experiments, while training set size was systematically varied according to the supervision scale defined in the experimental framework (section \ref{Experiment}).
\vspace{-8pt}
\subsection{Model architecture}
The study employs the Segment Anything Model (SAM3) \citep{carion2025sam3}, a foundation VLM model for promptable segmentation. The model supports multiple prompt modalities, including textual and geometric inputs, which are encoded into a shared embedding space through a unified perception encoder and fused with image features via cross-attention.

Object recognition and localization are decoupled via a dedicated presence head that first estimates the presence of the queried concept at the image level before performing instance-level localization. Localization is performed using a DETR-based architecture with learned object queries, enabling detection of multiple instances per concept. Pixel-level segmentation masks are generated by a MaskFormer-style decoder, which predicts object masks directly from query embeddings rather than via dense per-pixel classification.

The model's modular decomposition into semantic, spatial, and decoding components is leveraged here to enable selective adaptation during FT, in which different components are updated based on the prompting pipeline, as detailed in Section \ref{FT}.
\vspace{-8pt}
\subsection{Text prompt selection} \label{prompt_selection}
Selecting an effective semantic descriptor is a non-trivial task in SAM3, as the model lacks a built-in mechanism for identifying optimal prompts conditioned on the target domain. While VLMs capable of generating image captions (e.g., BLIP or Florence) could be used for this purpose, their reliance on natural-image training data limits their suitability for RS imagery, where object appearance and context differ substantially. Consequently, prompt selection in overhead imagery remains largely empirical. 

To address this, we adopt a structured multi-stage procedure for selecting an appropriate noun phrase for our target. Starting from an initial seed term, a diverse set of candidate noun phrases is generated using LLM-based paraphrasing, producing semantically related variants that capture different possible semantic descriptions of the target. The generated candidates are then aligned with the conceptual space of SAM3 by mapping them to corresponding entities in Wikidata \citep{vrandevcic2014}, which is used as the basis for the SA-Co ontology \citep{carion2025sam3}. This step ensures that only semantically valid and visually groundable concepts, consistent with model’s vocabulary, are retained. From this filtered set, a subset of $M=5$ candidate prompts is selected for empirical evaluation. The final set of prompts was chosen to represent a diverse set of ontology-consistent descriptions of the target concept while maintaining a tractable evaluation procedure. Rather than attempting an exhaustive search of the prompt space, the objective was to identify a robust semantic descriptor for subsequent experiments.

To prevent information leakage, prompt selection was performed using a dedicated dataset derived from a spatially distinct subset of the study area that was excluded from all subsequent experiments. The prompt-selection dataset followed the same sampling design as the primary dataset and included imagery from all acquisition years, ensuring that candidate prompts were evaluated across the full range of imaging conditions represented in the study. Each yearly subset consisted of 1,400 annotated tiles, with an average density of $\sim$1.8 PV installations per tile.

Let $P = \{p_1, \dots, p_M\}$ denote the resulting set of candidate prompts. Each candidate, $p_i$, is fine-tuned on training sets of 200 to 700 samples (in increments of 100) and evaluated using a held-out validation set of 200 samples and a test set of 500 samples. For each candidate, performance is summarized using a scalar score $S(p_i)$, defined as the mean of F1 and IoU averaged over the set of all evaluation settings ($\mathcal{C}$), (Eq. \ref{prompt_avg}). This formulation jointly captures detection accuracy and spatial overlap, ensuring a balanced evaluation of segmentation quality. Averaging performance across all supervision levels was intended to identify a prompt that remained robust across a broad range of adaptation settings. The goal was not to optimize performance for a particular supervision scale or imaging condition. The objective was therefore to select a prompt exhibiting consistent performance across experimental settings rather than one tailored to a specific configuration.

\begin{equation}
S_{p_i}=\frac{1}{|\mathcal{C}|}\sum_{c \in \mathcal{C}}
\frac{\mathrm{F1}_{i,c}+\mathrm{IoU}_{i,c}}{2}
\label{prompt_avg}
\end{equation}

The optimal prompt $p_{\text{opt}}$ is selected as the candidate with the highest score (Eq. \ref{best}):
\begin{equation}
p_{\text{opt}} = \arg\max_{p_i \in P} S(p_i)
\label{best}
\end{equation}

\subsection{Experimental design}\label{Experiment}
The experimental design systematically varies the prompting strategy, training regime, and supervision scale to assess their effects on segmentation performance for small-scale PV installations under varying imaging conditions. The study adopts a controlled within-model design.

\begin{figure*}[t]
    \centering
    \includegraphics[width=1\textwidth]{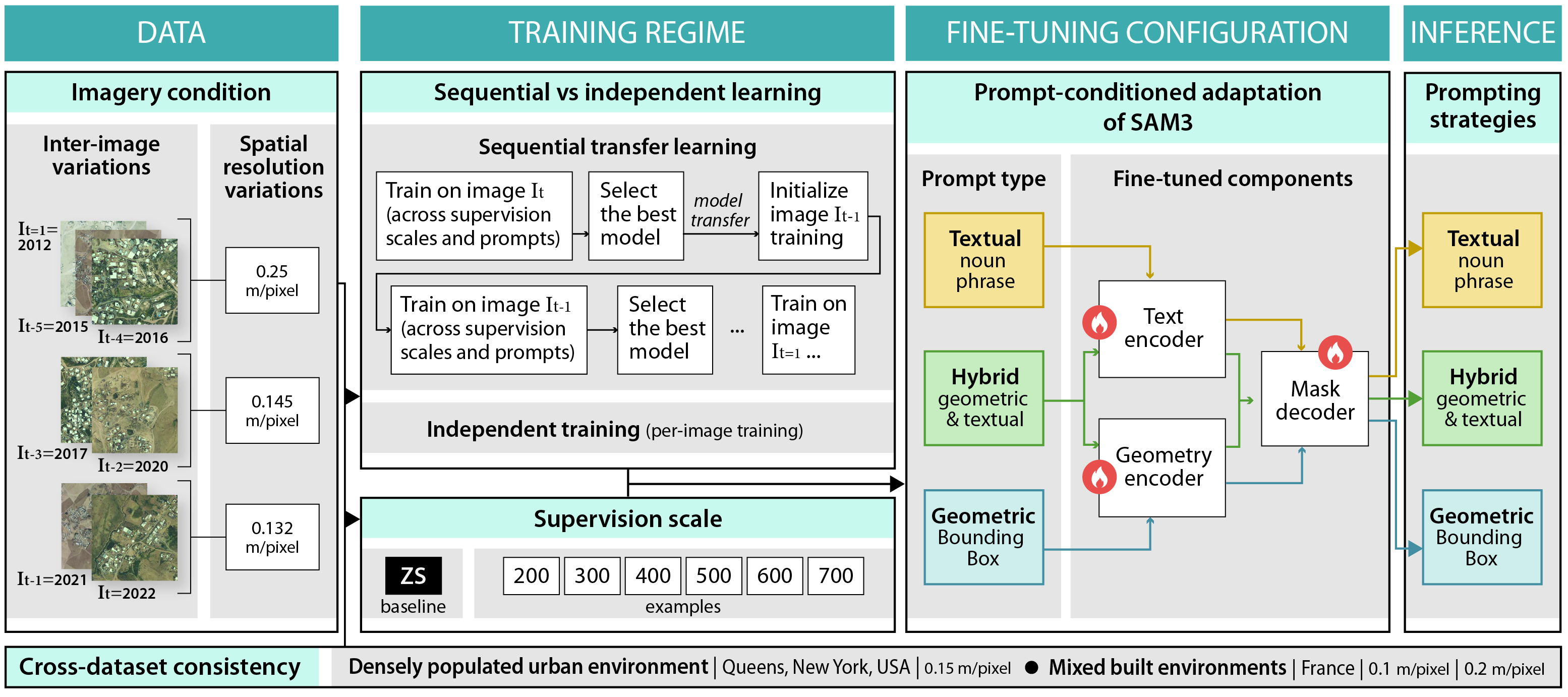}
    \vspace{-14pt}
    \caption{\textbf{Experimental framework.} SAM3 performance is systematically evaluated as a function of RS data, training regime, and inference prompting strategy. Data variation captures differences in imaging conditions. The training regime compares independent per-image training and sequential TL across varying supervision scales. Prompt-conditioned FT configurations define the prompt type and the model's unfrozen components. The corresponding prompting strategy generates the segmentation outputs. The analysis is further extended to additional datasets to assess result consistency.} 
    \label{framework}
\end{figure*}

\subsubsection{Prompting strategies}
Three prompting strategies are evaluated:
\begin{itemize}[leftmargin=*, itemsep=1pt, topsep=1pt, labelsep=6pt]
\item \textit{Textual prompting:} segmentation guided solely by natural language descriptions of PV installations, providing semantic guidance without explicit spatial cues (see Section \ref{prompt_selection} for prompt formulation).
\item \textit{Geometric prompting:} segmentation guided by BBs derived from ground-truth masks, providing explicit spatial guidance.
\item \textit{Hybrid prompting:} segmentation guided by a combination of textual descriptions and BBs, integrating both semantic and spatial guidance.
\end{itemize}
This comparison enables assessing the relative contribution of semantic and spatial guidance to small-scale PV segmentation and evaluating how model behavior changes when both forms of guidance are provided simultaneously.

Ground-truth-derived BBs were used as a controlled source of spatial guidance to isolate the contribution of prompting type. Using predicted detections would have confounded the effects of guidance modality with detector errors, preventing a controlled comparison between semantic and spatial guidance. Consequently, the present study focuses on a controlled evaluation of prompting strategies rather than on the performance of a fully automated segmentation pipeline.

\vspace{7pt}
\subsubsection{Training regime}\label{regime}
Training regime is designed to evaluate effects of both training strategy and supervision scale, under limited data availability. 

\textbf{\textit{Training strategy}}: Two complementary training strategies are implemented, alongside a systematic variation in supervision scale:
\begin{itemize}[leftmargin=*, itemsep=2pt, topsep=2pt, labelsep=6pt]
\item \textit{Independent training (per-image FT)}: Each image is treated as an independent task. The model is initialized from the pretrained SAM3 weights and fine-tuned separately for each image, without incorporating information from other images.
\item \textit{Sequential transfer learning (TL)}: A cumulative training scheme in which model parameters are progressively transferred across images, allowing each subsequent model to leverage knowledge acquired during previous training stages \citep{zhao2024comparison}. TL is conducted separately for each prompting pipeline (see \textit{Training configuration}), with knowledge transferred only within the corresponding prompting strategy. The TL process begins with the highest-resolution image (2022), on which models are trained across all supervision scales. The best-performing model is used to initialize training on the next image (2021), where it is first evaluated in ZS setting and subsequently fine-tuned across all supervision scales. This procedure is repeated iteratively, with each stage leveraging accumulated knowledge from the best-performing models in previous stages.
\end{itemize}

\vspace{2pt}
\textbf{\textit{Supervision scale}}: To assess data efficiency, the supervision scale is systematically varied, with training set sizes ranging from 200 to 700 samples in increments of 100. In addition, ZS performance is evaluated by applying the pretrained model without FT, providing a baseline for intrinsic generalization capability.

\vspace{7pt}
\subsubsection{imaging conditions}
The experimental design explicitly leverages variation in image characteristics at two levels: spatial resolution and cross-image variability within each resolution group.

\vspace{2pt}
\textbf{\textit{Spatial resolution}}: Three spatial resolution levels were defined (Table \ref{res_year}): 0.132, 0.145, and 0.25 m/pixel, all of which fall within the range considered suitable for detecting small-scale residential PV installations \citep{lu2024pv, jiang2021multi}, ensuring that observed performance differences reflect model sensitivity rather than resolution constraints. To assess the effect of spatial resolution on segmentation performance, we fitted a fractional logit model with F1-score as the response variable. This modeling framework is appropriate for bounded response variables defined on the unit interval and avoids assumptions of normally distributed and homoscedastic residuals. Prompting strategy, spatial resolution, and their interaction were included as explanatory variables. To isolate the effect of spatial resolution, training strategy and supervision scale were included as covariates to control for variation in training regime. Statistical significance was evaluated using robust Wald Type-II tests based on HC3 heteroscedasticity-consistent standard errors. Estimated marginal means were then computed for each prompt-resolution combination, and Holm-corrected pairwise comparisons were used to evaluate differences between resolution levels within each prompting strategy.

\vspace{2pt}
\textbf{\textit{Cross-image variability}}: To assess cross-image variability, performance differences among acquisition years were evaluated separately for each prompt-resolution combination after controlling for training size and training regime. Adjusted mean F1 values were estimated for each year, and year effects were evaluated using Type-II ANOVA. As the primary objective of this analysis was to quantify and compare the magnitude of year-related performance variability across prompt-resolution combinations, effect sizes were estimated using partial eta squared ($\eta_p^2$). Significant effects were further examined using Holm-corrected post-hoc comparisons.

To characterize radiometric differences among images acquired at the same spatial resolution, each pixel was represented by an \textit{intensity} value computed as the mean of its RGB channels. Because the imagery was acquired under heterogeneous imaging conditions and lacked consistent radiometric calibration information,  radiometric variability was characterized using image-based measures derived directly from pixel intensity distributions. Image \textit{contrast} was quantified using a percentile-based dynamic range ($DR=P_{95}-P_{5}$) derived from the distribution of pixel intensities, a robust measure that reduces sensitivity to extreme pixel values \citep{hulusic2017robust}. Additional percentile ranges ($P_{90}\text{-}P_{10}$, $P_{98}\text{-}P_{2}$, and $P_{99}\text{-}P_{1}$) were examined to verify the robustness of the contrast characterization and produced an identical ranking of imagery contrast.

\subsubsection{Cross-dataset consistency}
To evaluate the generality of the observed performance patterns across diverse geographic and environmental settings, experiments were further conducted on two publicly available aerial imagery datasets spanning three spatial resolution levels comparable to those examined in the primary study dataset. The first comprises rooftop PV installations across multiple regions in France and consists of two subsets: a Google Earth subset (0.1 m/pixel) and an IGN subset (0.2 m/pixel), representing a broad range of built environments and architectural characteristics \citep{kasmi2023}. The second consists of rooftop PV installations in Queens, New York, USA, acquired at approximately 0.15 m/pixel, representing a densely populated urban environment characterized by numerous large flat-roof buildings \citep{furedi2026}. 

\begin{table}[t]
\centering
\tablefont
\caption{\textbf{Characteristics of the additional benchmark datasets.}}
\label{tab:external_datasets}

\renewcommand{\arraystretch}{1.15}
\setlength{\tabcolsep}{2pt}

\begin{tabularx}{\linewidth}{@{}
>{\hsize=1.45\hsize\raggedright\arraybackslash}X
>{\hsize=0.95\hsize\centering\arraybackslash}X
>{\hsize=0.95\hsize\centering\arraybackslash}X
>{\hsize=0.65\hsize\centering\arraybackslash}X
>{\hsize=0.80\hsize\centering\arraybackslash}X
>{\hsize=1.00\hsize\centering\arraybackslash}X
@{}}

\toprule

\makecell[tl]{\textbf{Dataset}}
&
\makecell[tc]{%
\textbf{Spatial}\\
\textbf{resolution}\\
(m/pixel)}
&
\makecell[tc]{%
\textbf{Annotated}\\
\textbf{PVs}\\
(count)}
&
\makecell[tc]{%
\textbf{FG:}\\
\textbf{BG}\\
\textbf{ratio}}
&
\makecell[tc]{%
\textbf{Median}\\
\textbf{PV size}\\
(m\textsuperscript{2})}
&
\makecell[tc]{%
\textbf{PV size}\\
\textbf{IQR}\\
(m\textsuperscript{2})}
\\

\midrule

France (Google)
&
0.10
&
15,025
&
1:35
&
18.6
&
14.1-22.4
\\

France (IGN)
&
0.20
&
9,989
&
1:75
&
20.0
&
15.7-25.1
\\

Queens, NY
&
0.15
&
12,871
&
1:21
&
25.1
&
13.9-40.3
\\

\bottomrule
\end{tabularx}

\vspace{1mm}

\begin{minipage}{\linewidth}
\tablenotefont
\textbf{Note:} Statistics are computed from the complete benchmark datasets.
FG:BG ratio refers to the ratio of foreground to background pixels. 
\end{minipage}
\end{table}

To ensure consistency, each dataset was evaluated using the same prompting strategies, supervision scales (ZS and 200–700 samples), fixed validation (200 samples) and test (500 samples) sets, and evaluation metrics employed in the primary experiment.

To assess whether the relative performance of the prompting strategies remained consistent across datasets, we fitted a fractional logit model with F1-score as the response variable. Prompting strategy, dataset, and their interaction were included as explanatory variables. Supervision scale was included as a covariate to control for differences associated with training sample size. Statistical significance was evaluated using robust Wald Type-II tests based on HC3 heteroscedasticity-consistent standard errors. Estimated marginal means were then computed for each prompt–dataset combination, and Holm-corrected pairwise comparisons were used to compare prompting strategies within each dataset.
\vspace{-8pt}
\subsection{Fine-tuning configuration} \label{FT}
Leveraging the modular architecture of SAM3, fine-tuning (FT) is selectively applied to model components via complementary prompting pipelines that differ in the guidance provided to the model. To isolate the contribution of semantic adaptation, the textual-only pipeline relies exclusively on a selected conceptual noun phrase. In this setting, the text encoder is unfrozen during training, while the visual backbone and geometric components remain frozen. In the geometric-only pipeline, segmentation is guided solely by spatial cues (BBs), with the geometry encoder unfrozen during training while the text encoder remains frozen. In the geometric-textual pipeline, segmentation is guided by a combination of the conceptual noun phrase and spatial cues. Both the text encoder and the geometry encoder are unfrozen during training, enabling joint adaptation of semantic and spatial representations. In all pipelines, the mask decoder is unfrozen to adapt pixel-level predictions to the target domain, while the vision encoder and the DETR-based encoder–decoder remain frozen to preserve pretrained representations. To ensure comparability across experimental conditions, all training runs use identical hyperparameters (learning rate of $1\times10^{-4}$, 25 epochs, batch size of 2). 

All models were implemented in PyTorch using the Hugging Face Transformers library and fine-tuned from the SAM3 Large checkpoint on an NVIDIA RTX A4000 GPU (16 GB memory). Optimization was performed using Adam. Following the official SAM3 implementation, the training objective combined Dice loss and Focal loss to address severe class imbalance while promoting accurate mask delineation \citep{carion2025sam3}.
\vspace{-8pt}
\subsection{Evaluation metrics}
Given the severe class imbalance, evaluation focused on metrics that capture the trade-off between false positives and false negatives. Precision reflects the correctness of detected PV pixels, while Recall captures the completeness of the detected PV extent. The F1-score summarizes this balance, whereas IoU provides a stricter assessment of spatial overlap by quantifying the geometric agreement between predicted and ground-truth masks. Predicted probability maps were thresholded at 0.5 without additional post-processing.

\section{Results}
\subsection{Text prompt selection}
\begin{figure}[h!]
    \centering
    \fbox{\includegraphics[width=0.97\columnwidth]{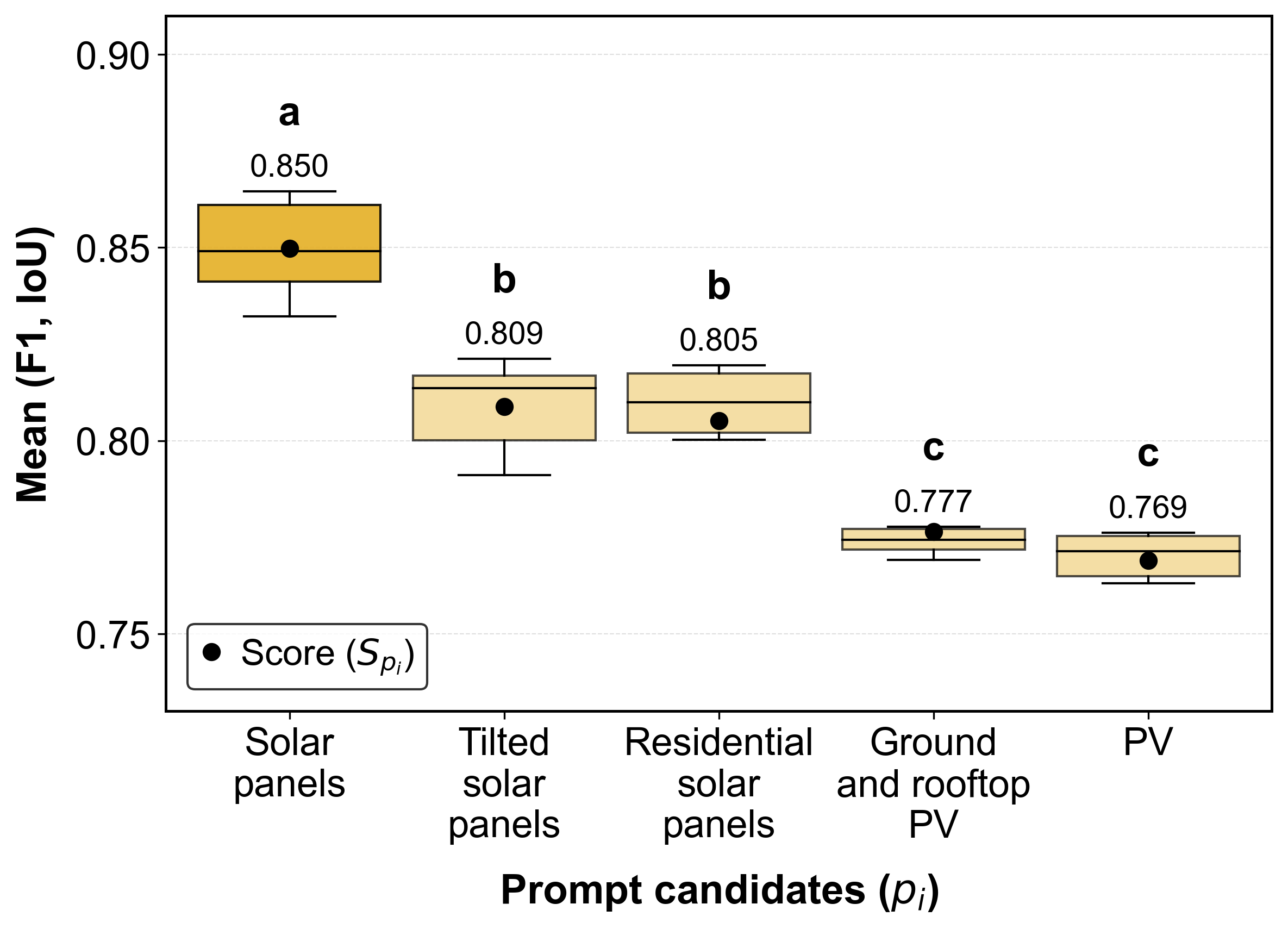}}
\caption{\textbf{Text prompt performance comparison.} Distribution of per-setting performance for each prompt $p_i$, measured as the mean of F1 and IoU. Boxplots summarize results across all evaluation settings ($\mathcal{C}$). Mean score values are annotated above each box. Different letters indicate statistically significant differences (Holm-corrected Wilcoxon tests, $p\!<\!0.05$).}
    \label{text_candidates}    
\end{figure}

Figure~\ref{text_candidates} reveals clear performance differences between candidate prompts. The prompt \textit{solar panels} achieves the highest overall score, $S(p_i)\!=\!0.85$, and it significantly outperforms all other candidates. Accordingly, it was selected as the optimal prompt $p_{\text{opt}}$.

\subsection{Prompting strategies}
\begin{figure}[h!]
    \centering
    \fbox{\includegraphics[width=0.97\columnwidth]{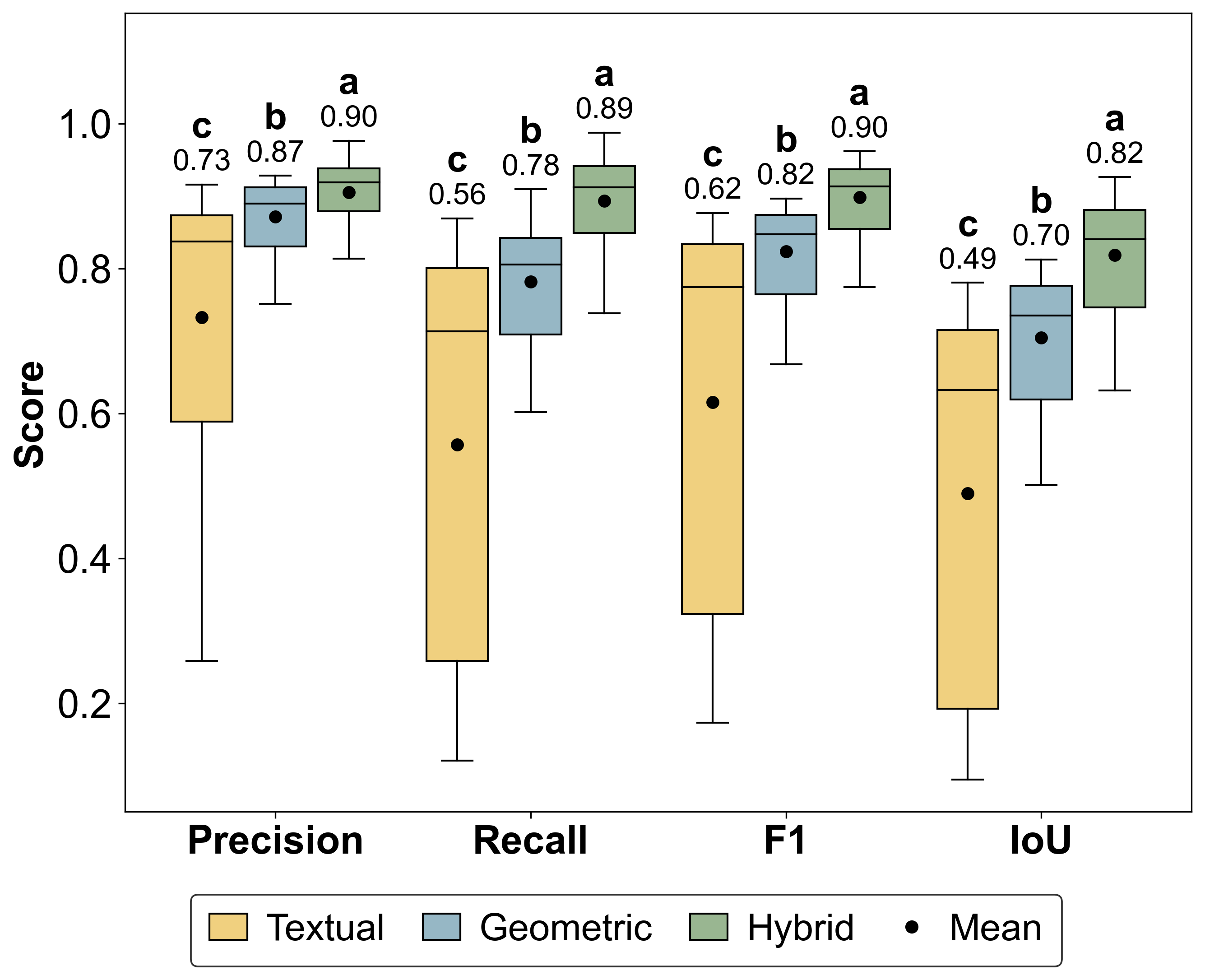}}
\caption{\textbf{Performance metrics by prompting strategy.} Boxplots summarize performance distributions across all experiments (excluding ZS). A clear and stable hierarchy is observed across all metrics, with hybrid prompting achieving the highest scores, followed by geometric and textual prompting. Strategies incorporating spatial guidance produce substantially more stable performance distributions. Mean score values are annotated above each box. Different letters denote statistically significant differences ($P<0.05$).}
    \label{prompt}    
\end{figure}

Figure~\ref{prompt} examines whether semantic guidance alone can support reliable segmentation of small-scale PV installations and how its performance compares with strategies incorporating explicit spatial information. Across all imaging conditions and training regimes (excluding ZS settings), textual prompting consistently yielded the lowest performance across all evaluation metrics. Incorporating spatial guidance substantially improved segmentation performance, with geometric prompting consistently outperforming textual prompting and hybrid prompting achieving the highest scores. Differences among prompting strategies were statistically significant for all metrics (Friedman repeated-measures tests followed by Holm-corrected pairwise Wilcoxon comparisons, $p\!<\!0.05$). Beyond improving average performance, spatial guidance substantially reduced performance variability across experimental settings. Textual prompting exhibited markedly broader score distributions, indicating unstable localization under heterogeneous imaging conditions. In contrast, geometric and hybrid prompting produced more compact distributions, suggesting improved robustness.

Replacing textual prompting with geometric prompting increased aggregated mean Recall and IoU by +0.22 and +0.21, respectively, while Precision improved more moderately (+0.14), indicating that textual prompting struggled more with detection completeness than with prediction correctness. These results highlight the critical role of explicit spatial cues for both localization and accurate segmentation. Nevertheless, incorporating semantic guidance alongside spatial cues further improved mean F1 by +0.08 relative to geometric prompting alone. This additional improvement was driven mainly by Recall (+0.11), whereas Precision increased only marginally (+0.03). 

\subsection{Training regime}
\subsubsection{Training strategy}
Figure \ref{regime} compares segmentation performance under the two training strategies, independent per-image training (No TL) and sequential transfer learning (TL), across supervision scales and prompting strategies. Overall, the impact of TL was strongly dependent on the prompting strategy, with meaningful benefits observed only when segmentation relied on semantic guidance.

The largest gains from TL were observed under textual prompting. While average ZS performance remained nearly unchanged, TL improved mean F1 by $\sim$ +0.08 at 200 training samples, with advantages persisting up to 400 samples before gradually converging at larger supervision scales. Under geometric prompting, TL produced a notable improvement in average ZS performance, increasing mean F1 by $\sim$+0.19 relative to independent inference. However, once supervision was introduced, performance rapidly increased under both training regimes and largely converged from 200 training samples onward. Hybrid prompting showed an intermediate pattern, exhibiting a modest improvement under ZS inference ($\sim$+0.07 F1), although this benefit was substantially smaller than that observed under geometric prompting and largely disappeared once FT data became available.
\vspace{5pt}
\subsubsection{Supervision scale}
Figure \ref{res} shows segmentation performance across supervision levels and spatial resolutions. Across all prompting strategies, Performance increased sharply with 200–300 training samples. However, performance trends differed substantially between prompting strategies incorporating spatial guidance and semantic-only guidance. Under geometric and hybrid prompting, performance reached a plateau after $\sim$300 samples. At this supervision level, hybrid prompting already achieved an average F1 score of $\sim$0.90 when aggregated across years and spatial resolutions. Textual prompting continued to improve up to $\sim$400 training samples, followed by a gradual decline in Recall and IoU at higher supervision scales, while Precision remained comparatively stable. The pattern was consistently observed across all spatial resolution levels, but was most pronounced at 0.25 m/pixel and became progressively weaker at higher resolutions. 

\begin{figure*}[h!]
\centering
\fbox{\includegraphics[width=0.99\textwidth]{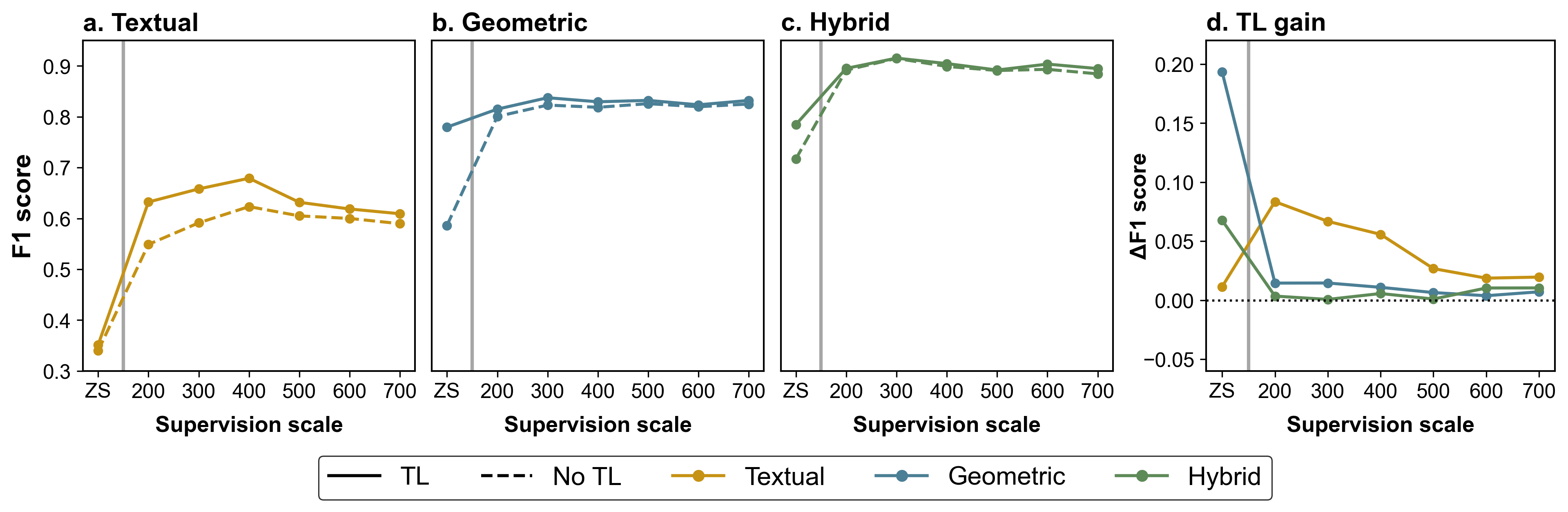}}
\caption{\textbf{Training regime effects.} Panels a–c compare TL and independent (No TL) training strategy across supervision scale and prompting type. Panel d shows the mean F1 gain of TL relative to No TL. TL mainly benefited textual prompting at low supervision, while under geometric and hybrid prompting its effects were largely limited to ZS, with larger gains for geometric prompting. Performance under geometric and hybrid prompting plateaued at 300 samples, whereas textual prompting peaked at 400 before declining.}
    \label{regime}    
\end{figure*} 

\subsection{imaging conditions}
\subsubsection{Spatial resolution}
\begin{figure}[h!]
    \centering
    \fbox{\includegraphics[width=0.95\columnwidth]{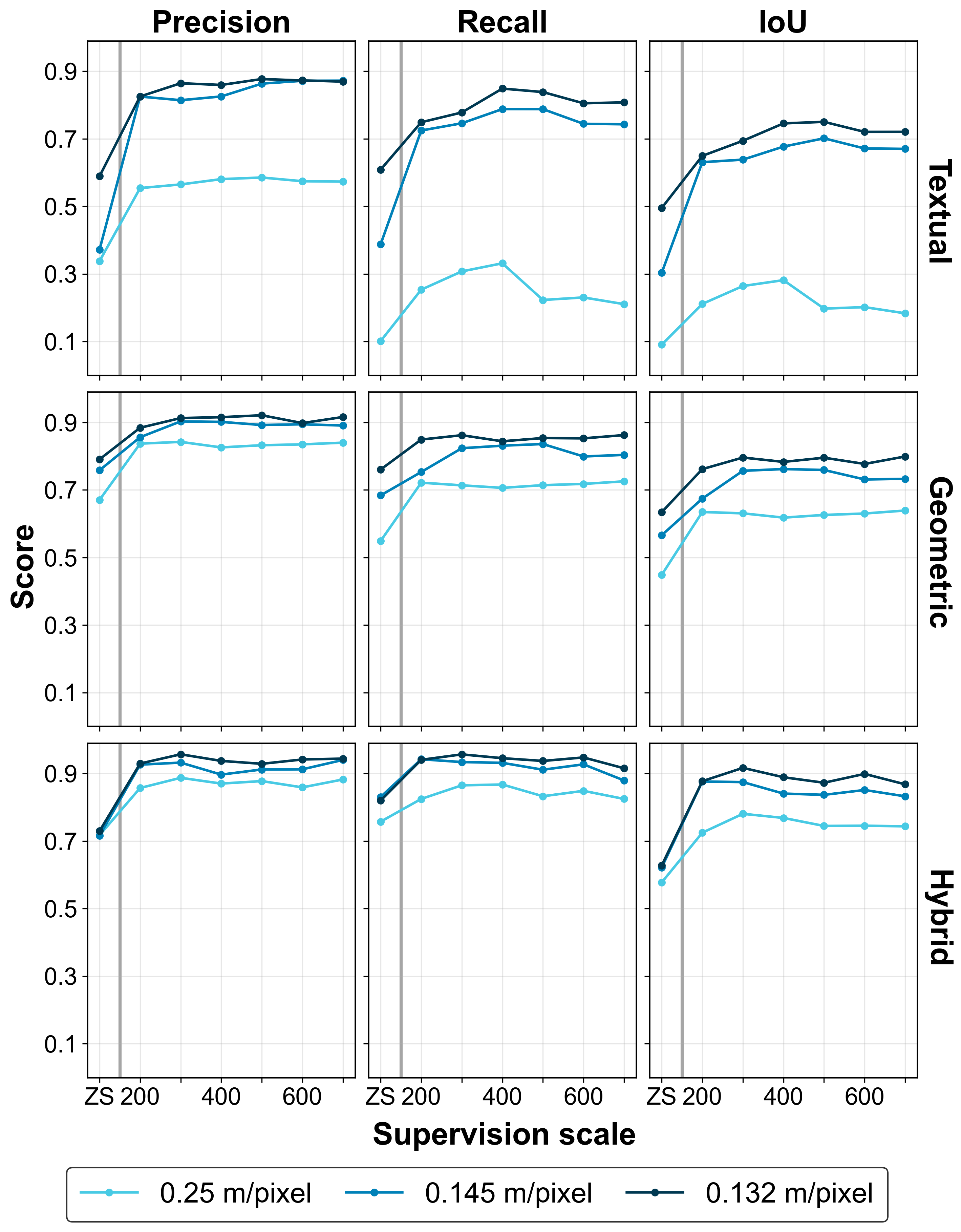}}
\caption{\textbf{Effect of spatial resolution across prompting strategies and supervision scales.} Higher resolution generally improved segmentation performance, with textual prompting showing the greatest sensitivity to resolution. Resolution effects were substantially smaller under geometric and hybrid prompting and were strongest for Recall and IoU.}
    \label{res}    
\end{figure}

Figure \ref{res} reveals a consistent positive effect of spatial resolution on segmentation performance across all prompting strategies. To formally assess these differences, we fitted a fractional logit model with F1-score as the response variable while controlling for training strategy and supervision scale. Robust Wald Type-II tests indicated highly significant effects of prompting strategy ($\chi^2\!=\!414.82$, $df\!=\!2$, $p\!<\!0.001$), spatial resolution ($\chi^2\!=\!360.54$, $df\!=\!2$, $p\!<\!0.001$), and their interaction ($\chi^2\!=\!144.81$, $df\!=\!4$, $p\!<\!0.001$), indicating that the influence of spatial resolution depended on prompting strategy.

Estimated marginal means revealed substantial differences in resolution sensitivity among prompting approaches. Under textual prompting, estimated F1 increased from 0.37 at 0.25 m/pixel to 0.81 at 0.145 m/pixel and 0.84 at 0.132 m/pixel. In contrast, geometric prompting increased more moderately from 0.78 to 0.86 and 0.89, while hybrid prompting increased from 0.87 to 0.92 and 0.94. These results indicate that textual prompting exhibited the strongest sensitivity to image resolution, whereas hybrid prompting remained comparatively stable, across resolution levels, while consistently achieving the highest overall performance.

Holm-corrected pairwise comparisons showed that increasing resolution from 0.25 to 0.145 m/pixel significantly improved performance across all prompting strategies ($p\!<\!0.001$). Further increases from 0.145 to 0.132 m/pixel also remained significant ($p\!<\!0.01$), but the associated performance gains were substantially smaller. These results indicate diminishing returns at the highest spatial resolution.

\begin{figure*}[h!]
    \centering
    \setlength{\fboxrule}{0.5pt}
    \setlength{\fboxsep}{3pt}
\fbox{\includegraphics[width=0.95\textwidth]{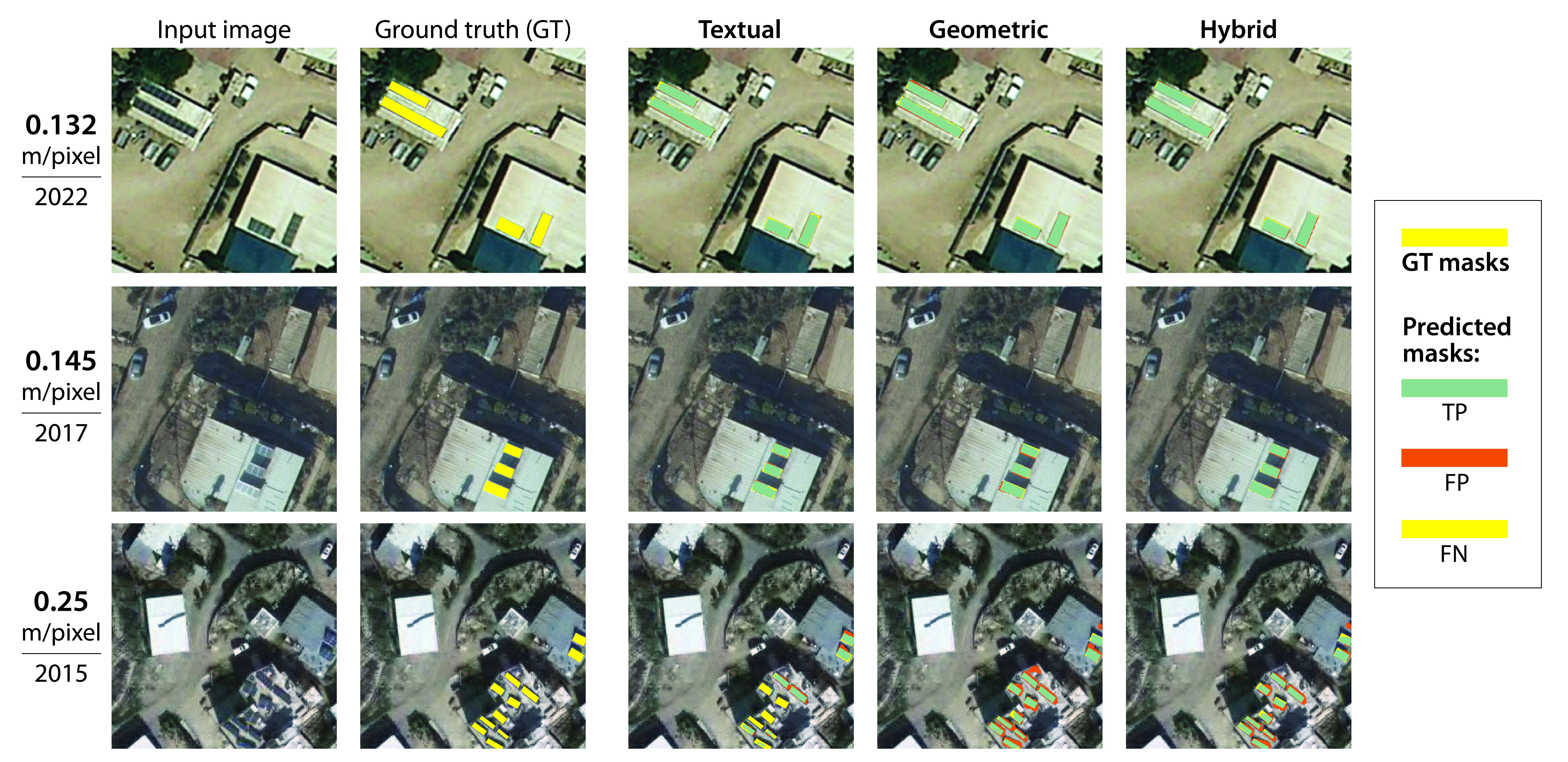}}
\caption{\textbf{Qualitative comparison of prompting strategies across spatial resolutions.} Representative examples of PV segmentation under textual, geometric, and hybrid prompting across three spatial resolutions. Predicted masks are decomposed into true positives (TP), false positives (FP), and false negatives (FN), highlighting differences in segmentation quality and error patterns. Predictions correspond with the best-performing model configuration (training regime and supervision level) for each prompting strategy and image.}
    \label{tiles}    
\end{figure*} 

Performance gaps between resolution levels were substantially smaller under geometric and hybrid prompting than under textual prompting, indicating that spatial guidance reduces dependence on image detail. Moreover, hybrid prompting consistently exhibited the smallest resolution-related performance differences and the highest overall performance across resolution levels. Representative examples in Figure \ref{tiles} further illustrate this behavior. While segmentation quality deteriorates visibly under textual prompting as resolution decreases, strategies incorporating spatial guidance maintain substantially better localization and boundary delineation across resolution levels.

Resolution effects also differed across evaluation metrics: they were most pronounced for Recall and IoU, whereas Precision showed smaller differences and converged more rapidly as supervision increased. This pattern suggests that higher resolution primarily improves object completeness and boundary delineation rather than reducing false-positive detections.

Despite these differences, effect of resolution remained largely consistent across supervision scales. Notably, under textual prompting at 0.25 m/pixel, Recall and IoU peaked at $\sim$400 training samples before declining at larger supervision scales, indicating reduced training stability under challenging image conditions.
\vspace{5pt}
\subsubsection{Cross-image variability}
Table~\ref{years} summarizes image contrast and cross-image performance variability across acquisition years within each spatial resolution group. Image contrast is represented by the percentile-based dynamic range (DR). Cross-image performance variability is quantified by the range of adjusted F1 values ($\Delta$F1) across acquisition years after controlling for training size and training regime, together with the corresponding effect size of acquisition year ($\eta_p^2$) estimated using Type-II ANOVA.

Patterns of year-related performance variability differed across prompting strategies and resolution groups, and generally reflected radiometric differences among imagery acquired at the same resolution. The largest contrast differences between images were observed at 0.25~m/pixel resolution ($\Delta\mathrm{DR}\!=\!87$), where acquisition year also had a substantial effect on F1 score. 

ANOVA results revealed that the effect of acquisition year on segmentation performance was strongest under textual prompting, where year accounted for $\approx$ 68\% of the explained variance ($\eta_p^2\!=\!0.679$), but was reduced by nearly half under geometric prompting ($\eta_p^2\!=\!0.356$) and by approximately two-thirds under hybrid prompting ($\eta_p^2\!=\!0.237$). Consistent with these effect sizes, cross-image differences in adjusted F1 decreased substantially with the incorporation of spatial guidance. In contrast, imagery acquired at 0.145~m/pixel exhibited virtually no radiometric variation ($\Delta\mathrm{DR}\!=\!1$) and also no significant performance differences between acquisition years under any prompting strategy. Although imagery acquired at 0.132~m/pixel displayed radiometric differences ($\Delta\mathrm{DR}\!=\!28$), these were not accompanied by significant performance differences under any prompting strategy.
Detailed Type-II ANOVA results and radiometric characteristics of the individual acquisition years are provided in supplementary table \ref{anova} and \ref{radiometry}, respectively.

\begin{table}[t]
\centering
\caption{\textbf{Cross-image variability across prompting strategies and spatial resolutions.}
Effect sizes were estimated using Type-II ANOVA while controlling for training size and training strategy. DR denotes the percentile-based dynamic range of the intensity distribution, used here as a measure of image contrast. $\Delta$F1 denotes the range of adjusted F1 values across acquisition years.}
\label{years}
\footnotesize
\renewcommand{\arraystretch}{1.15}
\setlength{\tabcolsep}{6pt}
\begin{tabular}{p{1.2cm}p{1.2cm}p{1.6cm}p{0.6cm}p{1.8cm}}
\toprule
\multicolumn{2}{c}{\small\textbf{Image characteristics}} &
\multicolumn{3}{l}{\hspace{1.5mm}\small\textbf{Cross-image effects on adjusted F1}} \\
\cmidrule(r{0.3em}){1-2}
\cmidrule(l{0.3em}){3-5}
\textbf{Resolution}
&
\hspace{2mm} \textbf{DR}
&
\hspace{1mm} \textbf{Prompt}
&
\textbf{$\Delta$F1}
&
\textbf{Effect size ($\eta_p^2$)} \\
\midrule
\hspace{2mm} \multirow{3}{*}{\makecell{\textbf{0.25}\\m/pixel}}
&
\multirow{3}{*}{
\makecell[l]{
Min=75\\
Max=162\\
$\Delta$=87}}
& \hspace{2mm} \textbf{Textual} & 0.278 & \hspace{3mm} 0.679 *** \\
&
& \hspace{2mm} \textbf{Geometric} & 0.097 & \hspace{3mm} 0.356*** \\
&
& \hspace{2mm} \textbf{Hybrid} & 0.052 & \hspace{3mm} 0.237* \\
\cmidrule{1-5}
\hspace{2mm} \multirow{3}{*}{\makecell{\textbf{0.145}\\ m/pixel}}
&
\multirow{3}{*}{
\makecell[l]{
Min=132\\
Max=133\\
$\Delta$=1}}
& \hspace{2mm} \textbf{Textual} & 0.063 & \hspace{3mm} 0.092 \\
&
& \hspace{2mm} \textbf{Geometric} & 0.044 & \hspace{3mm} 0.061  \\
&
& \hspace{2mm} \textbf{Hybrid} & 0.007 & \hspace{3mm} 0.010  \\
\cmidrule{1-5}
\hspace{2mm} \multirow{3}{*}{\makecell{\textbf{0.132}\\ m/pixel}}
&
\multirow{3}{*}{
\makecell[l]{
Min=130\\
Max=158\\
$\Delta$=28}}
& \hspace{2mm} \textbf{Textual} & 0.070 & \hspace{3mm} 0.113 \\
&
& \hspace{2mm} \textbf{Geometric} & 0.051 & \hspace{3mm} 0.098 \\
&
& \hspace{2mm} \textbf{Hybrid} & 0.010 & \hspace{3mm} 0.081 \\
\bottomrule
\end{tabular}
\vspace{-6pt} \begin{center} \footnotesize * $p < 0.05$; ** $p < 0.01$; *** $p < 0.001$. \end{center}
\end{table}

\subsubsection{Cross-dataset consistency}
\begin{figure}[h!]
    \centering
    \setlength{\fboxrule}{0.5pt}
    \setlength{\fboxsep}{3pt}
\fbox{\includegraphics[width=0.95\columnwidth]{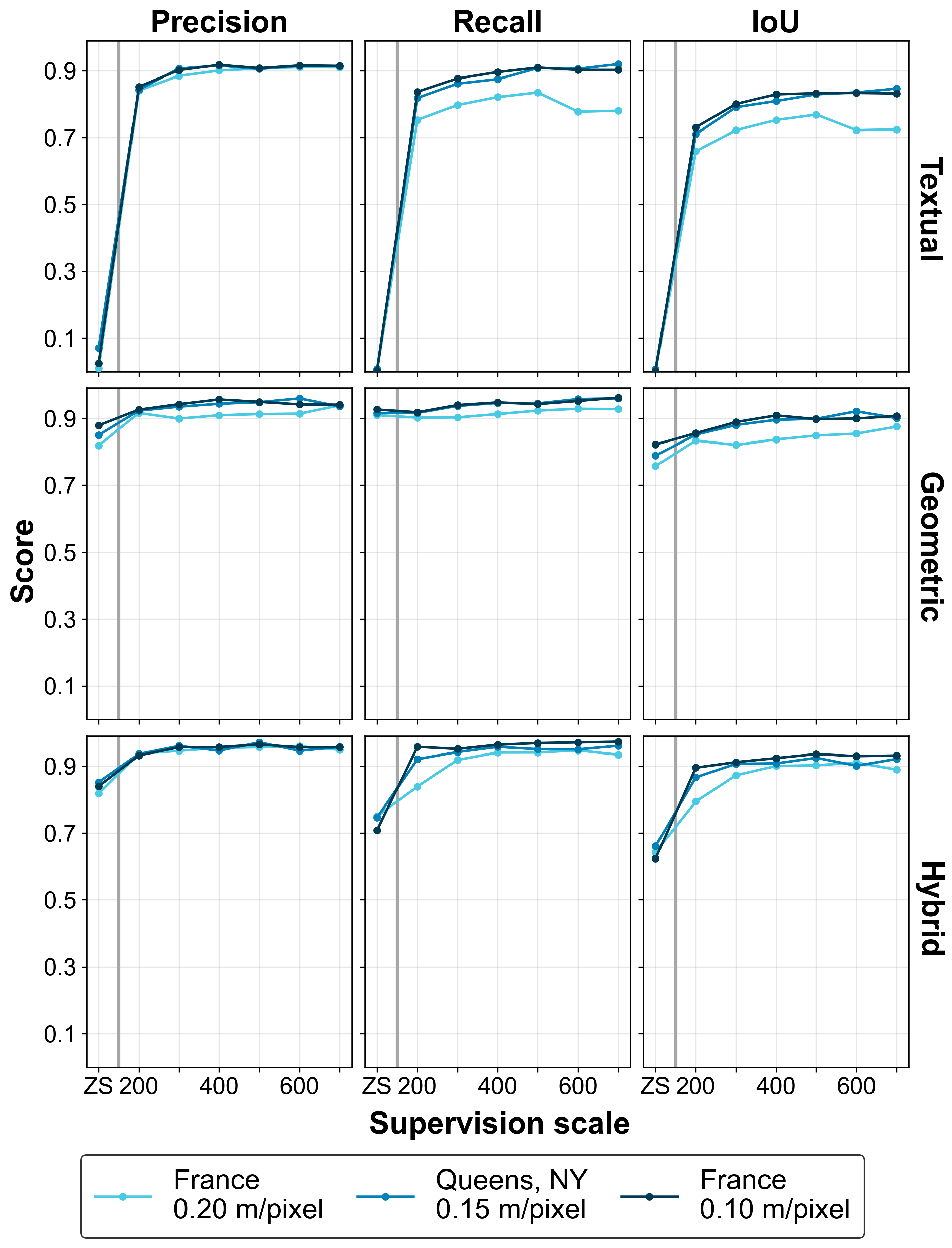}}
\caption{\textbf{Performance of prompting strategies across spatial resolutions and supervision scales in the benchmark datasets.}  The performance hierarchy observed in the primary dataset was preserved, with most gains achieved after the first 200-300 annotated samples. Overall performance was higher, whereas differences between spatial resolutions were smaller.}
    \label{dbs_res_sample}    
\end{figure}

Figure \ref{dbs_res_sample} presents segmentation performance across the three benchmark datasets as a function of prompting strategy and supervision scale. Consistent with the primary dataset, hybrid prompting outperformed geometric prompting, which in turn outperformed textual prompting across all evaluation metrics. Notably, overall performance was consistently higher than in the primary dataset, and performance differences between resolution levels were reduced across all prompting strategies, most notably under textual prompting.

As in the primary dataset, performance increased rapidly within the first few hundred annotated samples. Geometric and hybrid prompting plateaued after 200–300 samples, whereas textual prompting continued to improve until 500 samples before declining at 0.20 m/pixel in Recall and IoU, while Precision remained comparatively stable. The largest differences were observed for the 0.20 m/pixel dataset and were most pronounced in Recall and IoU under textual prompting. The 0.10 and 0.15 m/pixel datasets showed comparable performance across all prompting strategies.

To formally evaluate these patterns, a fractional logit model was fitted with prompting strategy, dataset, and their interaction as explanatory variables while controlling for supervision scale. Robust Wald Type II tests showed that prompting strategy was the dominant source of variation (Wald $\chi^2$=809.8), followed by supervision scale ($\chi^2$=108.4) and dataset ($\chi^2$ = 102.7), with all effects statistically significant ($p<0.001$). In contrast, the interaction between prompting strategy and dataset was substantially weaker ($\chi^2$=9.5), reached marginal statistical significance ($p=0.0497$), and suggests only limited evidence that the effect of dataset differed across prompting strategies.

\begin{figure}[h!]
    \centering
    \setlength{\fboxrule}{0.5pt}
    \setlength{\fboxsep}{3pt}
\fbox{\includegraphics[width=0.95\columnwidth]{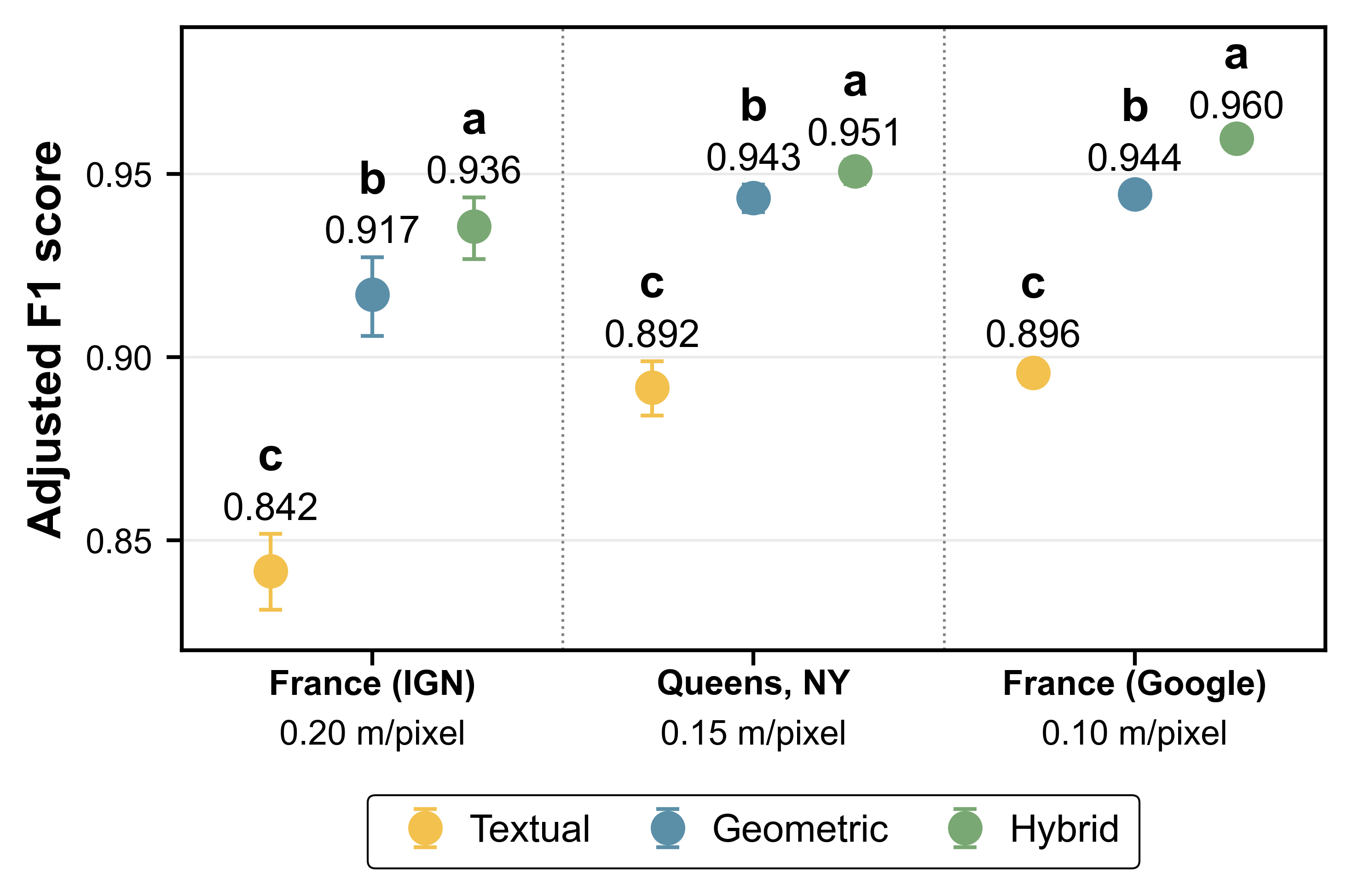}}
\caption{\textbf{EMMs of adjusted F1 scores by prompting strategy and dataset after controlling for supervision scale.} Error bars denote 95\% CI. Different letters denote significant differences within each dataset. The largest differences occurred between textual and the two spatially guided prompting strategies, particularly for the 0.20 m/pixel dataset.}
    \label{dbs_means}    
\end{figure}

Estimated marginal means (EMMs) of adjusted F1 scores (Figure \ref{dbs_means}) showed significant differences between all prompting strategies within each dataset, with all Holm-corrected pairwise comparisons remaining significant ($p<0.01$). The largest performance gaps between prompting strategies observed for the 0.20 m/pixel dataset and were driven primarily by the lower adjusted F1 of textual prompting. Confidence intervals were widest for this dataset across all prompting strategies.

\section{Discussion}
\subsection{Prompting strategies}
Direct comparisons of alternative prompting modalities remain rare in RS segmentation, leaving their role in shaping promptable FMs e.g., SAM3 unclear. We previously \citep{blushtein2026} revealed performance differences among prompting strategies, which motivated the current systematic evaluation across varying imaging and training conditions. We found here a clear performance hierarchy: textual prompting yielded the lowest accuracy; geometric prompting performed substantially better, and hybrid prompting consistently achieved the highest accuracy. This suggests that the two modalities are complementary rather than interchangeable, with textual prompts conveying target identity and geometric prompts providing localization cues.

Although VLMs achieve strong semantic alignment in RS imagery, translating semantic concepts into accurate pixel-level predictions remains difficult and their representations often lack the spatial granularity needed for precise delineation \citep{li2025segearth, zhou2024geo}. More broadly, language-guided segmentation requires aligning semantic concepts with pixel-level representations, a challenge recognized beyond RS \citep{wang2022cris}. Likewise, our findings suggest that under textual prompting, localization, rather than semantic recognition, is a major source of uncertainty for small PVs. The larger degradation in Recall than Precision supports this interpretation. Rather than confusing PVs with visually similar objects, the model primarily failed to recover all target pixels and instances, indicating limitations in spatial grounding rather than target identification.

Localization difficulties may be particularly pronounced for small, sparse, and weakly represented targets \citep{blushtein2026}. This is likely the case for small PVs in RS imagery, which occupy only a small fraction of the image, exhibit substantial visual variability, and are embedded within scenes dominated by non-target pixels. Thus, textual prompting requires the model to perform a global semantic search across large and heterogeneous images. Recent studies similarly identify numerous small objects, extreme scale variation, and visually similar targets as major challenges for language-guided RS segmentation \citep{li2025segearth, jiang2026, blushtein2026}.

Yet, the superior performance of hybrid prompting suggests that semantic guidance remains valuable even when spatial guidance is available. Although recent work shows that geometry-aware prompting improves PV segmentation by constraining the search space and encoding multi-scale structural information \citep{yao2026pvsam}, our results further indicate that spatial guidance primarily reduces localization uncertainty, whereas semantic guidance enriches the geometrically constrained search with higher-level knowledge about the target, improving performance, particularly Recall, by helping recover pixels that would otherwise be missed.  

Our analysis provides empirical support for recent developments in language-guided segmentation and visual grounding that increasingly integrate different modalities through approaches such as GeoGround, SegEarth-OV3, and GeoSAM \citep{zhou2024geo, li2025segearth, sultan2023geosam}. While these methods differ in implementation, they all combine semantic and spatial information, yet the individual contribution of each modality remains unresolved. Our systematic comparison disentangles these distinct and complementary contributions.

The hierarchy observed in the primary dataset was consistently reproduced across the external datasets, supporting the generality of prompting behavior. Moreover, comparison between the Queens dataset and the corresponding spatial-resolution subset of the primary dataset isolates the influence of target size and scene imbalance (Section~\ref{cross-dataset} and Supplementary table \ref{queens}). Despite their similar resolution, the Queens dataset, with its larger PVs and less severe foreground-background imbalance, achieved higher performance, particularly in Recall and IoU, with smaller differences between prompting strategies. These findings suggest that the contribution of semantic guidance to object delineation and pixel recovery diminishes as object size decreases and foreground-background imbalance increases.

To isolate the intrinsic contribution of spatial guidance from potential confounding effects of detector errors, we used ground-truth-derived BBs, thereby ensuring that performance differences could be attributed directly to the prompting strategy. This controlled design, however, raises the practical question of whether geometric prompts can be generated reliably under real-world conditions characterized by small objects and severe foreground-background imbalance. To address this, we conducted preliminary experiments using a YOLO26l detector \citep{jocher2026yolo26}, fine-tuned on 700 samples under the same data and imaging conditions examined throughout this study. The detector was evaluated on one representative acquisition year from each spatial resolution group using confidence and IoU thresholds of 0.5, achieving recall/precision values of 0.898/0.869 at 0.132 m/pixel, 0.850/0.831 at 0.145 m/pixel, and 0.793/0.789 at 0.25 m/pixel. These results demonstrate the feasibility of detector-generated geometric prompts, although their impact on downstream SAM3 segmentation remains to be evaluated in a fully automated end-to-end pipeline.

\subsection{Training regime}
\subsubsection{Training strategy}
Our results indicate that the effect of training strategy depends strongly on prompting type. While TL consistently improved performance under textual prompting, its contribution under geometric and hybrid prompting was largely confined to the ZS setting, with training-regime differences rapidly diminishing as supervision increased. This suggests that TL benefits depend not only on transferred prior knowledge, but also on the information guiding model adaptation and prediction. By exposing the model to diverse visual manifestations of the same target class across acquisition campaigns, spatial resolutions, and imaging conditions, TL may have strengthened the alignment between semantic concepts and their visual representations, facilitating more effective semantic grounding during segmentation.

Recent developments in language-guided RS segmentation support this by emphasizing effective semantic-visual alignment and diverse textual representations, as reflected in frameworks such as RSPrompter and RemoteSAM \citep{chen2024rsprompter, yao2025remotesam}. We further suggest that TL reinforced semantic-visual alignment through a complementary mechanism by exposing the model to diverse visual manifestations of the same target across heterogeneous imaging conditions, rather than increasing textual representation variation.

The substantially smaller contribution of TL, once explicit spatial guidance was introduced, suggests a different underlying mechanism under geometric and hybrid prompting. We propose that explicit spatial cues reduce task ambiguity, limiting the extent to which prior knowledge acquired through TL can further improve performance. This interpretation is supported by the ZS results, where TL yielded its largest gains when explicit geometric information was available. Consequently, TL value may depend not only on the knowledge transferred from previous training stages, but also on the ambiguity remaining in the segmentation task.

\subsubsection{Supervision scale}
Performance saturated rapidly across all prompting strategies, with most gains achieved within the first 200 training samples. Similar patterns were reported for SAM adaptations in medical and microscopy domains. For example, both H-SAM and PTSAM demonstrated high sample efficiency, while H-SAM further showed only marginal gains from substantially increasing the amount of training data \citep{cheng2024hsam, piater2025prompt}. Within the RS domain, PointSAM \citep{liu2025pointsam} and Multiscale-SAM \citep{chen2025multiscale} demonstrated effective SAM adaptation using weak supervision and parameter-efficient FT, but provided limited insight into how performance scales with supervision. We address this gap by explicitly characterizing this relationship, providing empirical evidence for the sample efficiency of FM when segmenting small, highly imbalanced targets.

Prompting strategies also differed in the supervision required to reach peak performance. While geometric and hybrid prompting saturated at 300 training samples, textual prompting continued to improve until $\sim$400 samples. This delayed saturation likely reflects the additional learning burden of language-guided segmentation, where the model must learn both target semantics and their spatial grounding. Consistent with previous work identifying text-to-pixel alignment as a major challenge \citep{wang2022cris}, the additional supervision required by textual prompting may therefore reflect the need to learn cross-modal grounding in addition to segmentation itself.

At higher supervision levels, a different pattern emerged. Geometric and hybrid prompting remained stable, whereas textual prompting exhibited a decline in Recall and IoU beyond 400 training samples, while Precision remained comparatively stable. This selective pattern may indicate reduced sensitivity to more challenging or atypical PV installations rather than a general deterioration in performance. One possible explanation is that continued optimization progressively biases text-conditioned representations toward the dominant visual manifestations of PVs in the training distribution, reducing sensitivity to less common appearances at inference. Under this interpretation, the initial increase in supervision improves domain adaptation and semantic grounding, whereas continued optimization provides progressively less new information while increasingly reinforcing the dominant visual patterns already learned. Consistent with this interpretation, the effect was observed across all spatial resolution levels but became progressively weaker as resolution increased, suggesting that greater image detail partially mitigates this tendency.

Although the proposed mechanism cannot be directly verified from the present experiments, related phenomena have been reported across a range of FMs and VLMs, suggesting that continued adaptation may increase specialization at the expense of robustness. For example, \citet{kumar2022} and \citet{wortsman2022} showed that FT can distort pretrained representations and reduce out-of-distribution generalization, potentially narrowing the diversity of visual patterns captured by the model. Similarly, \citet{you2024} reported that multimodal adaptation may improve performance for dominant groups while reducing robustness for atypical or underrepresented examples, and \citet{labonte2024} further showed that continued optimization can degrade minority-group performance even when overall performance remains stable. Given the substantial heterogeneity of PV installations in the present dataset, a similar effect may contribute to the decline in Recall under textual prompting. Nonetheless, further research is needed to determine whether this behavior reflects over-specialization of text-conditioned representations or another form of adaptation instability.

\subsection{Spatial resolution}
Spatial resolution consistently improved segmentation performance across all prompting strategies, with the strongest effects on Recall and IoU, indicating that higher resolution primarily enhanced localization and boundary delineation rather than target identification. This interpretation is supported by recent PV segmentation studies, which increasingly identify localization and boundary delineation as key challenges and address them through edge-aware learning, scale-adaptive representations, and boundary-refinement mechanisms \citep{guo2024transpv, li2025joint, wang2025pv}. While these studies focus on architectural solutions, our results suggest that image resolution itself is also a key determinant of localization and delineation quality.

Perhaps the most important finding is that resolution sensitivity depended strongly on the prompting strategy. The highly significant interaction between spatial resolution and prompting indicates that image quality and prompting strategy jointly shape segmentation performance rather than acting independently. Textual prompting exhibited the greatest sensitivity to resolution, whereas geometric and especially hybrid prompting were considerably more robust. Building on the distinct roles of semantic and spatial guidance discussed above, these findings suggest that semantic guidance benefits disproportionately from increased spatial detail, whereas spatial guidance can partially compensate for reduced image quality. Consequently, our results extend previous work on language-guided segmentation by indicating that, in SAM3-based PV segmentation, the effectiveness of semantic grounding is itself conditioned by image resolution.

The superior stability of hybrid prompting provides an additional perspective on robustness under heterogeneous imaging conditions. Across all resolutions, it consistently achieved both the highest performance and the smallest performance differences, indicating that combining semantic and spatial guidance reduces dependence on fine visual detail. This robustness highlights the potential of hybrid prompting for operational mapping applications where imagery originates from multiple acquisition campaigns and image characteristics cannot be fully controlled. More broadly, the results suggest that the benefits of increasingly higher spatial resolution diminish once sufficient object detail is available, particularly when segmentation is supported by complementary prompting strategies.

Another notable finding concerns the non-linear relationship between spatial resolution and segmentation performance. Most gains were achieved when resolution increased from 0.25 to 0.145 m/pixel, whereas the further increase to 0.132 m/pixel yielded significantly smaller improvements despite remaining statistically significant. Our results extend those of \citet{li2021understanding}. Whereas they reported little performance difference between 0.30 and 0.15 m imagery, our results indicate that further improvements remain achievable at higher resolutions, although with progressively diminishing returns. A straightforward explanation for this non-linear pattern lies in the number of pixels representing the target. In our dataset, median-sized PV systems were represented by approximately 69, 205, and 248 pixels at the three resolutions, meaning that the largest performance improvement coincided with an approximately threefold increase in pixel coverage, whereas the subsequent increase was comparatively modest. This interpretation is consistent with \citet{lu2024pv}, who emphasized that preserving fine spatial detail, including edge and texture information, is critical for small PVs segmentation. Increasing target pixel coverage naturally provides richer spatial detail, although the informational value of additional pixels appears to diminish once sufficient spatial detail has been captured. We suggest here that much of the benefit of higher spatial resolution may already be achieved once targets are represented by a few hundred pixels.

\subsection{Cross-image variability}
Cross-image variability affected segmentation performance primarily at the lowest spatial resolution and was substantially more pronounced under textual than geometric or hybrid prompting. Namely, sensitivity to imaging conditions depends not only on image quality but also on the guidance available for segmentation. Recent benchmark studies show that acquisition-related variability remains a fundamental challenge for Earth observation FMs, with performance degrading under changes in illumination, contrast, haze, blur, and noise \citep{li2026reobench}. Similar sensitivity to temporal changes in object appearance has also been reported in RS segmentation \citep{popp2023}. Our results extend these observations by showing that, in SAM3-based PV segmentation, sensitivity to acquisition-related variability is not determined solely by model architecture or image conditions, but is also strongly moderated by the prompting strategy itself.

Text-only prompting was particularly vulnerable to acquisition-related variability, likely because segmentation depended on image-wide semantic inference. Adding spatial guidance substantially reduced performance fluctuations across acquisition years, with hybrid prompting exhibiting the smallest cross-image variability. These findings suggest that spatial guidance serves not only as a localization cue but also as a stabilizing mechanism under heterogeneous imaging conditions.

Another notable observation is that the influence of cross-image variability was not consistent across spatial resolutions. Whereas substantial year effects were observed at 0.25 m/pixel, they disappeared entirely at 0.132 m/pixel despite measurable radiometric differences among acquisition years. This suggests that the influence of radiometric variability depends on the amount of spatial detail available, with higher resolution providing richer object representation that preserves accurate localization and delineation despite differences in image appearance. Although the relative contributions of spatial resolution and radiometric variability remain to be disentangled, these findings suggest that high-resolution imagery improves not only segmentation accuracy but also robustness to acquisition-related variability.
\vspace{-8pt}
\subsection{Cross-dataset consistency} \label{cross-dataset}
Evaluation on the external benchmark datasets demonstrated that the main findings generalized beyond the primary study area. The same prompting hierarchy was consistently reproduced, with hybrid prompting outperforming geometric prompting and textual prompting yielding the lowest performance. Similar training dynamics were also observed, with most performance gains achieved using only a few hundred annotated samples. Likewise, textual prompting remained the most sensitive to spatial resolution, with resolution effects most pronounced for Recall and IoU, whereas geometric and hybrid prompting remained comparatively stable. These findings suggest that the relationships between prompting strategy, supervision, and imaging conditions are robust across diverse RS datasets.

Despite these consistent trends, overall segmentation performance was uniformly higher in the benchmark datasets. A likely explanation is that they represent substantially less challenging segmentation conditions owing to both reduced class imbalance and substantially larger PV installations. Foreground-background ratios ranged from 1:21 to 1:75, compared with 1:207 in the primary dataset, corresponding to a 3- to 10-fold reduction in class imbalance. Likewise, median PV size ranged from 18.6 to 25.1 m², compared with 4.33 m² in the primary dataset. Notably, the lower quartile of PV size in the benchmark datasets (13.9-15.7 m²) exceeded the upper quartile of the primary dataset (6.76 m²), underscoring the substantial shift in target size and pixel representation. Reduced class imbalance likely facilitates target localization and delineation, particularly under textual prompting. Larger PV installations, in turn, provide richer visual information and greater pixel representation per object. These characteristics likely reduce the influence of spatial resolution, as targets remain sufficiently resolved even at 0.20 m/pixel.

To further examine this interpretation while minimizing the effect of spatial resolution, we compared the two datasets with the most similar resolution: the primary dataset at 0.145 m/pixel (2017 and 2020 imagery) and the Queens benchmark dataset at 0.15 m/pixel. Mean performance was averaged across all supervision levels using independently trained models only (Supplementary Table~\ref{queens}). Under both textual and geometric prompting, performance improvements were driven primarily by larger gains in Recall and IoU than in Precision. Notably, geometric prompting showed improvements comparable to textual prompting despite receiving explicit spatial guidance, indicating that the superior performance of the benchmark dataset cannot be explained by easier target localization alone. Rather, reduced class imbalance and larger PVs appear to improve mask delineation and target-pixel recovery. By contrast, hybrid prompting showed only modest improvements across all metrics, suggesting that it was already operating close to its performance ceiling under the more challenging conditions of the primary dataset.
\vspace{-8pt}
\subsection{Limitations and future research}
Despite these contributions, several limitations should be acknowledged. \textit{First}, the analysis focused on a single target category, namely small-scale PV installations. Although these represent a particularly challenging segmentation task, whether the observed relationships generalize to other object categories remains unknown. Future studies should therefore evaluate additional small targets with different spatial and visual characteristics. \textit{Second}, only a single FM architecture, SAM3, was evaluated. Assessing alternative promptable FMs would help determine the extent to which the findings generalize across architectures. \textit{Third}, geometric and hybrid prompting relied on ground-truth-derived BBs, which enabled a controlled evaluation of prompting strategies but provided idealized spatial guidance unavailable in fully automated settings. Although preliminary detector experiments demonstrated the feasibility of automatically generating BBs under the evaluated data and imaging conditions, the effect of detector-generated prompts on downstream SAM3 segmentation has yet to be evaluated. Future work should therefore assess fully automated detection-segmentation pipelines incorporating detector-generated spatial guidance. \textit{Finally}, radiometric variability was characterized using image-based measures derived from RGB intensity distributions as radiometric calibration data were not consistently available across acquisition campaigns. Radiometrically calibrated imagery could help disentangle the effects of specific acquisition-related factors, including illumination, atmospheric conditions, and sensor characteristics.

\section{Conclusions}
Our findings lead to the following five conclusions:
\begin{enumerate}[label=\arabic*., leftmargin=*, labelsep=0.5em, itemsep=2pt, topsep=2pt]
\item Within the evaluated experimental conditions, prompting strategy emerged as a primary factor governing model behavior. Beyond influencing segmentation accuracy, prompting systematically altered the model’s sensitivity to supervision scale, transfer learning, spatial resolution, and cross-image variability, resulting in markedly different performance patterns within the same FM. These findings suggest that for SAM3-based PV segmentation, prompting should be viewed not merely as an inference mechanism, but as a central design component shaping model adaptation, robustness, and generalization. In particular, the results underscore the importance of guidance type as a key consideration when adapting promptable segmentation FMs such as SAM3 for challenging RS tasks involving small PV installations and severe foreground-background imbalance. Whether similar relationships hold across other FM architectures and segmentation targets has yet to be established.
\item Textual prompting consistently exhibited the highest sensitivity to both training regime and imaging conditions, deriving the greatest benefit from transfer learning, requiring more supervision to reach peak performance, and remaining the most vulnerable to variations in spatial resolution and image characteristics. These findings suggest that semantic guidance alone places substantial demands on both semantic grounding and precise localization in PV segmentation.
\item Under controlled conditions in which ground-truth-derived spatial guidance was available, strategies incorporating spatial information substantially improved both segmentation performance and robustness. Geometric prompting reduced sensitivity to spatial resolution and cross-image variability, highlighting the value of explicit localization cues for robust segmentation across heterogeneous acquisition conditions. However, accurate segmentation of very small PV installations also depends on accurate target delineation and complete target-pixel recovery, even when spatial guidance is available.
\item Within the evaluated experimental framework, hybrid prompting consistently achieved the highest segmentation performance while exhibiting the lowest sensitivity to changes in supervision scale, training strategy, spatial resolution, and cross-image variability. Our findings indicate that combining semantic recognition with explicit spatial constraints provides the most robust segmentation strategy under heterogeneous training and imaging conditions.
\item Our results demonstrate strong data efficiency, with most performance gains achieved using only a few hundred annotated samples. This finding highlights the potential of promptable FMs such as SAM3 to support accurate data curation and mapping of small, sparsely distributed targets in data-constrained environments. By reducing supervision requirements for accurate PV segmentation, such models may help overcome a key bottleneck in constructing fine-grained spatio-temporal datasets of PV installations, thereby supporting systematic empirical investigation of PV adoption processes in off-grid and underserved regions where reliable installation records are often unavailable.
\end{enumerate}

\section*{Acknowledgment}
This work was supported by Israel Science Foundation (ISF) under Grant 299/23.

\bibliographystyle{cas-model2-names}
\bibliography{references}

\begin{thebibliography}{74}
\expandafter\ifx\csname natexlab\endcsname\relax\def\natexlab#1{#1}\fi
\providecommand{\url}[1]{\texttt{#1}}
\providecommand{\href}[2]{#2}
\providecommand{\path}[1]{#1}
\providecommand{\DOIprefix}{doi:}
\providecommand{\ArXivprefix}{arXiv:}
\providecommand{\URLprefix}{URL: }
\providecommand{\Pubmedprefix}{pmid:}
\providecommand{\doi}[1]{\href{http://dx.doi.org/#1}{\path{#1}}}
\providecommand{\Pubmed}[1]{\href{pmid:#1}{\path{#1}}}
\providecommand{\bibinfo}[2]{#2}
\ifx\xfnm\relax \def\xfnm[#1]{\unskip,\space#1}\fi
\bibitem[{Adib et~al.(2025)Adib, Islam, Abid and Ahshan}]{adib2025deep}
\bibinfo{author}{Adib, A.U.R.}, \bibinfo{author}{Islam, M.}, \bibinfo{author}{Abid, M.S.}, \bibinfo{author}{Ahshan, R.}, \bibinfo{year}{2025}.
\newblock \bibinfo{title}{A deep learning based framework for solar panel segmentation and fault classification enhanced with explainable ai}.
\newblock \bibinfo{journal}{Solar Energy} \bibinfo{volume}{302}, \bibinfo{pages}{114058}.
\bibitem[{Alayrac et~al.(2022)Alayrac, Donahue, Luc, Miech, Barr, Hasson, Lenc, Mensch, Millican, Reynolds et~al.}]{alayrac2022flamingo}
\bibinfo{author}{Alayrac, J.B.}, \bibinfo{author}{Donahue, J.}, \bibinfo{author}{Luc, P.}, \bibinfo{author}{Miech, A.}, \bibinfo{author}{Barr, I.}, \bibinfo{author}{Hasson, Y.}, \bibinfo{author}{Lenc, K.}, \bibinfo{author}{Mensch, A.}, \bibinfo{author}{Millican, K.}, \bibinfo{author}{Reynolds, M.}, et~al., \bibinfo{year}{2022}.
\newblock \bibinfo{title}{Flamingo: a visual language model for few-shot learning}.
\newblock \bibinfo{journal}{Advances in neural information processing systems} \bibinfo{volume}{35}, \bibinfo{pages}{23716--23736}.
\bibitem[{Balta-Ozkan et~al.(2021)Balta-Ozkan, Yildirim, Connor, Truckell and Hart}]{BaltaOzkan2021}
\bibinfo{author}{Balta-Ozkan, N.}, \bibinfo{author}{Yildirim, J.}, \bibinfo{author}{Connor, P.M.}, \bibinfo{author}{Truckell, I.}, \bibinfo{author}{Hart, P.}, \bibinfo{year}{2021}.
\newblock \bibinfo{title}{Energy transition at local level: Analyzing the role of peer effects and socio-economic factors on uk solar photovoltaic deployment}.
\newblock \bibinfo{journal}{Energy Policy} \bibinfo{volume}{148}, \bibinfo{pages}{112004}.
\newblock \DOIprefix\doi{10.1016/j.enpol.2020.112004}.
\bibitem[{Bluestein-Livnon et~al.(2023)Bluestein-Livnon, Svoray, Dorman and Van Der~Beek}]{bluestein2023economic}
\bibinfo{author}{Bluestein-Livnon, R.}, \bibinfo{author}{Svoray, T.}, \bibinfo{author}{Dorman, M.}, \bibinfo{author}{Van Der~Beek, K.}, \bibinfo{year}{2023}.
\newblock \bibinfo{title}{Economic aspects of urban greenness along a dryland rainfall gradient: A time-series analysis}.
\newblock \bibinfo{journal}{Urban Forestry \& Urban Greening} \bibinfo{volume}{83}, \bibinfo{pages}{127915}.
\bibitem[{Blushtein-Livnon et~al.(2026)Blushtein-Livnon, Rafaeli, Ioffe, Boger, Esquenazi and Svoray}]{blushtein2026}
\bibinfo{author}{Blushtein-Livnon, R.}, \bibinfo{author}{Rafaeli, O.}, \bibinfo{author}{Ioffe, D.}, \bibinfo{author}{Boger, A.}, \bibinfo{author}{Esquenazi, K.S.}, \bibinfo{author}{Svoray, T.}, \bibinfo{year}{2026}.
\newblock \bibinfo{title}{On the effectiveness of textual prompting with lightweight fine-tuning for sam3 remote sensing segmentation}.
\newblock \bibinfo{journal}{IEEE Geoscience and Remote Sensing Letters} .
\bibitem[{Blushtein-Livnon et~al.(2025)Blushtein-Livnon, Svoray and Dorman}]{blushtein2025performance}
\bibinfo{author}{Blushtein-Livnon, R.}, \bibinfo{author}{Svoray, T.}, \bibinfo{author}{Dorman, M.}, \bibinfo{year}{2025}.
\newblock \bibinfo{title}{Performance of human annotators in object detection and segmentation of remotely sensed data}.
\newblock \bibinfo{journal}{IEEE Transactions on Geoscience and Remote Sensing} \DOIprefix\doi{10.1109/TGRS.2025.3555235}.
\bibitem[{Camilo et~al.(2018)Camilo, Wang, Collins, Bradbury and Malof}]{camilo2018application}
\bibinfo{author}{Camilo, J.}, \bibinfo{author}{Wang, R.}, \bibinfo{author}{Collins, L.M.}, \bibinfo{author}{Bradbury, K.}, \bibinfo{author}{Malof, J.M.}, \bibinfo{year}{2018}.
\newblock \bibinfo{title}{Application of a semantic segmentation convolutional neural network for accurate automatic detection and mapping of solar photovoltaic arrays in aerial imagery}.
\newblock \bibinfo{journal}{arXiv preprint:1801.04018} .
\bibitem[{Carion et~al.(2025)Carion, Gustafson, Hu, Debnath, Hu et~al.}]{carion2025sam3}
\bibinfo{author}{Carion, N.}, \bibinfo{author}{Gustafson, L.}, \bibinfo{author}{Hu, Y.T.}, \bibinfo{author}{Debnath, S.}, \bibinfo{author}{Hu, R.}, et~al., \bibinfo{year}{2025}.
\newblock \bibinfo{title}{Sam 3: Segment anything with concepts}.
\newblock \bibinfo{journal}{arXiv preprint:2511.16719} .
\bibitem[{Chen et~al.(2024)Chen, Liu, Chen, Zhang, Li, Zou and Shi}]{chen2024rsprompter}
\bibinfo{author}{Chen, K.}, \bibinfo{author}{Liu, C.}, \bibinfo{author}{Chen, H.}, \bibinfo{author}{Zhang, H.}, \bibinfo{author}{Li, W.}, \bibinfo{author}{Zou, Z.}, \bibinfo{author}{Shi, Z.}, \bibinfo{year}{2024}.
\newblock \bibinfo{title}{Rsprompter: Learning to prompt for remote sensing instance segmentation based on visual foundation model}.
\newblock \bibinfo{journal}{IEEE Transactions on Geoscience and Remote Sensing} \bibinfo{volume}{62}, \bibinfo{pages}{1--17}.
\bibitem[{Chen et~al.(2025a)Chen, Yu, Li, Wang, Li and Han}]{chen2025multiscale}
\bibinfo{author}{Chen, S.}, \bibinfo{author}{Yu, Y.}, \bibinfo{author}{Li, Y.}, \bibinfo{author}{Wang, Z.}, \bibinfo{author}{Li, X.}, \bibinfo{author}{Han, J.}, \bibinfo{year}{2025}a.
\newblock \bibinfo{title}{Multiscale adapter based on sam for remote sensing semantic segmentation}.
\newblock \bibinfo{journal}{IEEE Journal of Selected Topics in Applied Earth Observations and Remote Sensing} \bibinfo{volume}{18}, \bibinfo{pages}{6806--6819}.
\bibitem[{Chen et~al.(2025b)Chen, Deng, Jin, Chen, Chen, Feng, Xi, Liu, Li and Meng}]{chen2025dgtrsd}
\bibinfo{author}{Chen, W.}, \bibinfo{author}{Deng, Y.}, \bibinfo{author}{Jin, W.}, \bibinfo{author}{Chen, J.}, \bibinfo{author}{Chen, J.}, \bibinfo{author}{Feng, Y.}, \bibinfo{author}{Xi, Z.}, \bibinfo{author}{Liu, D.}, \bibinfo{author}{Li, K.}, \bibinfo{author}{Meng, Y.}, \bibinfo{year}{2025}b.
\newblock \bibinfo{title}{Dgtrsd and dgtrsclip: A dual-granularity remote sensing image--text dataset and vision--language foundation model for alignment}.
\newblock \bibinfo{journal}{IEEE Journal of Selected Topics in Applied Earth Observations and Remote Sensing} \bibinfo{volume}{18}, \bibinfo{pages}{29113--29130}.
\bibitem[{Chen et~al.(2025c)Chen, Zhou, Chen, Wang, Zhang, Ge and Ma}]{chen2025edge}
\bibinfo{author}{Chen, Y.}, \bibinfo{author}{Zhou, J.}, \bibinfo{author}{Chen, Y.}, \bibinfo{author}{Wang, J.}, \bibinfo{author}{Zhang, X.}, \bibinfo{author}{Ge, Y.}, \bibinfo{author}{Ma, H.}, \bibinfo{year}{2025}c.
\newblock \bibinfo{title}{Edge-enhanced sam for extracting photovoltaic power plants from remote sensing imagery}.
\newblock \bibinfo{journal}{International Journal of Applied Earth Observation and Geoinformation} \bibinfo{volume}{140}, \bibinfo{pages}{104580}.
\bibitem[{Cheng et~al.(2022)Cheng, Misra, Schwing, Kirillov and Girdhar}]{cheng2022masked}
\bibinfo{author}{Cheng, B.}, \bibinfo{author}{Misra, I.}, \bibinfo{author}{Schwing, A.G.}, \bibinfo{author}{Kirillov, A.}, \bibinfo{author}{Girdhar, R.}, \bibinfo{year}{2022}.
\newblock \bibinfo{title}{Masked-attention mask transformer for universal image segmentation}, in: \bibinfo{booktitle}{Proceedings of the IEEE/CVF conference on computer vision and pattern recognition}, pp. \bibinfo{pages}{1290--1299}.
\bibitem[{Cheng et~al.(2024)Cheng, Wei, Zhu, Wang, Qu, Shao and Zhou}]{cheng2024hsam}
\bibinfo{author}{Cheng, Z.}, \bibinfo{author}{Wei, Q.}, \bibinfo{author}{Zhu, H.}, \bibinfo{author}{Wang, Y.}, \bibinfo{author}{Qu, L.}, \bibinfo{author}{Shao, W.}, \bibinfo{author}{Zhou, Y.}, \bibinfo{year}{2024}.
\newblock \bibinfo{title}{Unleashing the potential of sam for medical adaptation via hierarchical decoding}, in: \bibinfo{booktitle}{Proceedings of the IEEE/CVF conference on computer vision and pattern recognition}, pp. \bibinfo{pages}{3511--3522}.
\bibitem[{De~Groote et~al.(2016)De~Groote, Pepermans and Verboven}]{de2016heterogeneity}
\bibinfo{author}{De~Groote, O.}, \bibinfo{author}{Pepermans, G.}, \bibinfo{author}{Verboven, F.}, \bibinfo{year}{2016}.
\newblock \bibinfo{title}{Heterogeneity in the adoption of photovoltaic systems in flanders}.
\newblock \bibinfo{journal}{Energy economics} \bibinfo{volume}{59}, \bibinfo{pages}{45--57}.
\bibitem[{Diab et~al.(2025)Diab, Kolokoussis and Brovelli}]{diab2025optimizing}
\bibinfo{author}{Diab, M.}, \bibinfo{author}{Kolokoussis, P.}, \bibinfo{author}{Brovelli, M.A.}, \bibinfo{year}{2025}.
\newblock \bibinfo{title}{Optimizing zero-shot text-based segmentation of remote sensing imagery using sam and grounding dino}.
\newblock \bibinfo{journal}{Artificial Intelligence in Geosciences} \bibinfo{volume}{6}, \bibinfo{pages}{100105}.
\bibitem[{Furedi et~al.(2026)Furedi, Kimsal, Cornejo, Liero and Ranalli}]{furedi2026}
\bibinfo{author}{Furedi, T.}, \bibinfo{author}{Kimsal, E.}, \bibinfo{author}{Cornejo, S.}, \bibinfo{author}{Liero, N.}, \bibinfo{author}{Ranalli, J.}, \bibinfo{year}{2026}.
\newblock \bibinfo{title}{Labeled photovoltaic installations for orthographic aerial imagery in queens, new york}.
\newblock \bibinfo{journal}{Scientific Data} \bibinfo{volume}{13}, \bibinfo{pages}{207}.
\newblock \DOIprefix\doi{10.1038/s41597-025-06523-2}.
\bibitem[{Garc{\'\i}a et~al.(2024)Garc{\'\i}a, Aparcedo, Nayak, Ahmed, Shah and Li}]{garcia2024generalized}
\bibinfo{author}{Garc{\'\i}a, G.}, \bibinfo{author}{Aparcedo, A.}, \bibinfo{author}{Nayak, G.K.}, \bibinfo{author}{Ahmed, T.}, \bibinfo{author}{Shah, M.}, \bibinfo{author}{Li, M.}, \bibinfo{year}{2024}.
\newblock \bibinfo{title}{Generalized deep learning model for photovoltaic module segmentation from satellite and aerial imagery}.
\newblock \bibinfo{journal}{Solar Energy} \bibinfo{volume}{274}, \bibinfo{pages}{112539}.
\bibitem[{{GOGLA}(2023)}]{GOGLA2023}
\bibinfo{author}{{GOGLA}}, \bibinfo{year}{2023}.
\newblock \bibinfo{title}{Global Off-Grid Solar Market Report: Sales and Impact Data (H2 2022)}.
\newblock \bibinfo{type}{Technical Report}. GOGLA. \bibinfo{address}{Utrecht, The Netherlands}.
\newblock \URLprefix \url{https://www.gogla.org}. \bibinfo{note}{accessed 20 July 2025}.
\bibitem[{Graziano and Gillingham(2015)}]{Graziano2015}
\bibinfo{author}{Graziano, M.}, \bibinfo{author}{Gillingham, K.}, \bibinfo{year}{2015}.
\newblock \bibinfo{title}{Spatial patterns of solar photovoltaic system adoption: The influence of neighbors and the built environment}.
\newblock \bibinfo{journal}{Journal of Economic Geography} \bibinfo{volume}{15}, \bibinfo{pages}{815--839}.
\newblock \DOIprefix\doi{10.1093/jeg/lbu036}.
\bibitem[{Guo et~al.(2024)Guo, Lu, Chen, Liu, Song, Tan, Zhang and Yan}]{guo2024transpv}
\bibinfo{author}{Guo, Z.}, \bibinfo{author}{Lu, J.}, \bibinfo{author}{Chen, Q.}, \bibinfo{author}{Liu, Z.}, \bibinfo{author}{Song, C.}, \bibinfo{author}{Tan, H.}, \bibinfo{author}{Zhang, H.}, \bibinfo{author}{Yan, J.}, \bibinfo{year}{2024}.
\newblock \bibinfo{title}{Transpv: Refining photovoltaic panel detection accuracy through a vision transformer-based deep learning model}.
\newblock \bibinfo{journal}{Applied energy} \bibinfo{volume}{355}, \bibinfo{pages}{122282}.
\bibitem[{Hou et~al.(2019)Hou, Wang, Hu, Yin and Wu}]{hou2019solarnet}
\bibinfo{author}{Hou, X.}, \bibinfo{author}{Wang, B.}, \bibinfo{author}{Hu, W.}, \bibinfo{author}{Yin, L.}, \bibinfo{author}{Wu, H.}, \bibinfo{year}{2019}.
\newblock \bibinfo{title}{Solarnet: a deep learning framework to map solar power plants in china from satellite imagery}.
\newblock \bibinfo{journal}{arXiv preprint:1912.03685} .
\bibitem[{Hulusic et~al.(2017)Hulusic, Valenzise, Debattista and Dufaux}]{hulusic2017robust}
\bibinfo{author}{Hulusic, V.}, \bibinfo{author}{Valenzise, G.}, \bibinfo{author}{Debattista, K.}, \bibinfo{author}{Dufaux, F.}, \bibinfo{year}{2017}.
\newblock \bibinfo{title}{Robust dynamic range computation for high dynamic range content}, in: \bibinfo{booktitle}{IS\&T International Symposium on Electronic Imaging}.
\bibitem[{Huo et~al.(2025)Huo, Chen, Zhang, Wang, Yan, Shen, Hong, Qi, Fang and Wang}]{huo2025remote}
\bibinfo{author}{Huo, C.}, \bibinfo{author}{Chen, K.}, \bibinfo{author}{Zhang, S.}, \bibinfo{author}{Wang, Z.}, \bibinfo{author}{Yan, H.}, \bibinfo{author}{Shen, J.}, \bibinfo{author}{Hong, Y.}, \bibinfo{author}{Qi, G.}, \bibinfo{author}{Fang, H.}, \bibinfo{author}{Wang, Z.}, \bibinfo{year}{2025}.
\newblock \bibinfo{title}{When remote sensing meets foundation model: A survey and beyond}.
\newblock \bibinfo{journal}{remote sensing} \bibinfo{volume}{17}, \bibinfo{pages}{179}.
\bibitem[{{IEA}(2025)}]{IEA2025_population}
\bibinfo{author}{{IEA}}, \bibinfo{year}{2025}.
\newblock \bibinfo{title}{Population without electricity access, 2010--2025}.
\newblock \bibinfo{howpublished}{\url{https://www.iea.org/data-and-statistics/charts/population-without-electricity-access-2010-2025}}.
\newblock \bibinfo{note}{IEA, Paris}.
\bibitem[{IRENA et~al.(2025)IRENA, IEA, UNSD, Bank and WHO}]{IRENA_2025}
\bibinfo{author}{IRENA}, \bibinfo{author}{IEA}, \bibinfo{author}{UNSD}, \bibinfo{author}{Bank, W.}, \bibinfo{author}{WHO}, \bibinfo{year}{2025}.
\newblock \bibinfo{title}{Tracking SDG 7: The Energy Progress Report}.
\newblock \bibinfo{type}{Technical Report}. IRENA.
\newblock \URLprefix \url{https://trackingsdg7.esmap.org}.
\bibitem[{Jiang et~al.(2021)Jiang, Yao, Lu, Qin, Liu, Liu and Zhou}]{jiang2021multi}
\bibinfo{author}{Jiang, H.}, \bibinfo{author}{Yao, L.}, \bibinfo{author}{Lu, N.}, \bibinfo{author}{Qin, J.}, \bibinfo{author}{Liu, T.}, \bibinfo{author}{Liu, Y.}, \bibinfo{author}{Zhou, C.}, \bibinfo{year}{2021}.
\newblock \bibinfo{title}{Multi-resolution dataset for photovoltaic panel segmentation from satellite and aerial imagery}.
\newblock \bibinfo{journal}{Earth System Science Data Discussions} \bibinfo{volume}{2021}, \bibinfo{pages}{1--17}.
\bibitem[{Jiang et~al.(2026)Jiang, Pei, Zhao, Wu, Yu, Zhang, Cai et~al.}]{jiang2026}
\bibinfo{author}{Jiang, L.}, \bibinfo{author}{Pei, Y.}, \bibinfo{author}{Zhao, Y.}, \bibinfo{author}{Wu, T.}, \bibinfo{author}{Yu, S.}, \bibinfo{author}{Zhang, L.}, \bibinfo{author}{Cai, D.}, et~al., \bibinfo{year}{2026}.
\newblock \bibinfo{title}{Geoseg: Training-free reasoning-driven segmentation in remote sensing imagery}.
\newblock \bibinfo{journal}{arXiv preprint:2603.03983} .
\bibitem[{Jocher et~al.(2026)Jocher, Qiu, Liu, Lyu, Akyon and Kalfaoglu}]{jocher2026yolo26}
\bibinfo{author}{Jocher, G.}, \bibinfo{author}{Qiu, J.}, \bibinfo{author}{Liu, M.}, \bibinfo{author}{Lyu, S.}, \bibinfo{author}{Akyon, F.C.}, \bibinfo{author}{Kalfaoglu, M.E.}, \bibinfo{year}{2026}.
\newblock \bibinfo{title}{Ultralytics yolo26: Unified real-time end-to-end vision models}.
\newblock \URLprefix \url{https://arxiv.org/abs/2606.03748}, \DOIprefix\doi{10.48550/arXiv.2606.03748}, \href{http://arxiv.org/abs/2606.03748}{\tt arXiv:2606.03748}.
\bibitem[{Kasmi et~al.(2023)Kasmi, Saint-Drenan, Trebosc, Jolivet, Leloux, Sarr and Dubus}]{kasmi2023}
\bibinfo{author}{Kasmi, G.}, \bibinfo{author}{Saint-Drenan, Y.M.}, \bibinfo{author}{Trebosc, D.}, \bibinfo{author}{Jolivet, R.}, \bibinfo{author}{Leloux, J.}, \bibinfo{author}{Sarr, B.}, \bibinfo{author}{Dubus, L.}, \bibinfo{year}{2023}.
\newblock \bibinfo{title}{A crowdsourced dataset of aerial images with annotated solar photovoltaic arrays and installation metadata}.
\newblock \bibinfo{journal}{Scientific Data} \bibinfo{volume}{10}, \bibinfo{pages}{59}.
\newblock \DOIprefix\doi{10.1038/s41597-023-01951-4}.
\bibitem[{Kausika et~al.(2021)Kausika, Nijmeijer, Reimerink, Brouwer and Liem}]{kausika2021geoai}
\bibinfo{author}{Kausika, B.B.}, \bibinfo{author}{Nijmeijer, D.}, \bibinfo{author}{Reimerink, I.}, \bibinfo{author}{Brouwer, P.}, \bibinfo{author}{Liem, V.}, \bibinfo{year}{2021}.
\newblock \bibinfo{title}{Geoai for detection of solar photovoltaic installations in the netherlands}.
\newblock \bibinfo{journal}{Energy and AI} \bibinfo{volume}{6}, \bibinfo{pages}{100111}.
\bibitem[{Kirillov et~al.(2023)Kirillov, Mintun, Ravi, Mao, Rolland, Gustafson, Xiao, Whitehead, Berg, Lo et~al.}]{kirillov2023}
\bibinfo{author}{Kirillov, A.}, \bibinfo{author}{Mintun, E.}, \bibinfo{author}{Ravi, N.}, \bibinfo{author}{Mao, H.}, \bibinfo{author}{Rolland, C.}, \bibinfo{author}{Gustafson, L.}, \bibinfo{author}{Xiao, T.}, \bibinfo{author}{Whitehead, S.}, \bibinfo{author}{Berg, A.C.}, \bibinfo{author}{Lo, W.Y.}, et~al., \bibinfo{year}{2023}.
\newblock \bibinfo{title}{Segment anything}, in: \bibinfo{booktitle}{Proceedings of the IEEE/CVF international conference on computer vision}, pp. \bibinfo{pages}{4015--4026}.
\bibitem[{Kumar et~al.(2022)Kumar, Raghunathan, Jones, Ma and Liang}]{kumar2022}
\bibinfo{author}{Kumar, A.}, \bibinfo{author}{Raghunathan, A.}, \bibinfo{author}{Jones, R.}, \bibinfo{author}{Ma, T.}, \bibinfo{author}{Liang, P.}, \bibinfo{year}{2022}.
\newblock \bibinfo{title}{Fine-tuning can distort pretrained features and underperform out-of-distribution}.
\newblock \bibinfo{journal}{arXiv preprint:2202.10054} .
\bibitem[{LaBonte et~al.(2024)LaBonte, Hill, Zhang, Muthukumar and Kumar}]{labonte2024}
\bibinfo{author}{LaBonte, T.}, \bibinfo{author}{Hill, J.C.}, \bibinfo{author}{Zhang, X.}, \bibinfo{author}{Muthukumar, V.}, \bibinfo{author}{Kumar, A.}, \bibinfo{year}{2024}.
\newblock \bibinfo{title}{The group robustness is in the details: Revisiting finetuning under spurious correlations}.
\newblock \bibinfo{journal}{Advances in Neural Information Processing Systems} \bibinfo{volume}{37}, \bibinfo{pages}{121598--121629}.
\bibitem[{Li et~al.(2024a)Li, Cai, Li, Kou and Zhang}]{li2024review}
\bibinfo{author}{Li, J.}, \bibinfo{author}{Cai, Y.}, \bibinfo{author}{Li, Q.}, \bibinfo{author}{Kou, M.}, \bibinfo{author}{Zhang, T.}, \bibinfo{year}{2024}a.
\newblock \bibinfo{title}{A review of remote sensing image segmentation by deep learning methods}.
\newblock \bibinfo{journal}{International Journal of Digital Earth} \bibinfo{volume}{17}, \bibinfo{pages}{2328827}.
\bibitem[{Li et~al.(2025a)Li, Liu, Cao, Bai, Zhou, Meng and Wang}]{li2025segearth}
\bibinfo{author}{Li, K.}, \bibinfo{author}{Liu, R.}, \bibinfo{author}{Cao, X.}, \bibinfo{author}{Bai, X.}, \bibinfo{author}{Zhou, F.}, \bibinfo{author}{Meng, D.}, \bibinfo{author}{Wang, Z.}, \bibinfo{year}{2025}a.
\newblock \bibinfo{title}{Segearth-ov: Towards training-free open-vocabulary segmentation for remote sensing images}, in: \bibinfo{booktitle}{Proceedings of the Computer Vision and Pattern Recognition Conference}, pp. \bibinfo{pages}{10545--10556}.
\bibitem[{Li et~al.(2025b)Li, Lu and Qin}]{li2025joint}
\bibinfo{author}{Li, L.}, \bibinfo{author}{Lu, N.}, \bibinfo{author}{Qin, J.}, \bibinfo{year}{2025}b.
\newblock \bibinfo{title}{Joint-task learning framework with scale adaptive and position guidance modules for improved household rooftop photovoltaic segmentation in remote sensing image}.
\newblock \bibinfo{journal}{Applied Energy} \bibinfo{volume}{377}, \bibinfo{pages}{124521}.
\bibitem[{Li et~al.(2021)Li, Zhang, Guo, Lyu, Chen, Li, Song, Shibasaki and Yan}]{li2021understanding}
\bibinfo{author}{Li, P.}, \bibinfo{author}{Zhang, H.}, \bibinfo{author}{Guo, Z.}, \bibinfo{author}{Lyu, S.}, \bibinfo{author}{Chen, J.}, \bibinfo{author}{Li, W.}, \bibinfo{author}{Song, X.}, \bibinfo{author}{Shibasaki, R.}, \bibinfo{author}{Yan, J.}, \bibinfo{year}{2021}.
\newblock \bibinfo{title}{Understanding rooftop pv panel semantic segmentation of satellite and aerial images for better using machine learning}.
\newblock \bibinfo{journal}{Advances in applied energy} \bibinfo{volume}{4}, \bibinfo{pages}{100057}.
\bibitem[{Li et~al.(2026)Li, Tao, Zhang, Liu, Xiong, Luo, Liu, Pechenizkiy, Zhu and Huang}]{li2026reobench}
\bibinfo{author}{Li, X.}, \bibinfo{author}{Tao, Y.}, \bibinfo{author}{Zhang, S.}, \bibinfo{author}{Liu, S.}, \bibinfo{author}{Xiong, Z.}, \bibinfo{author}{Luo, C.}, \bibinfo{author}{Liu, L.}, \bibinfo{author}{Pechenizkiy, M.}, \bibinfo{author}{Zhu, X.}, \bibinfo{author}{Huang, T.}, \bibinfo{year}{2026}.
\newblock \bibinfo{title}{Reobench: Benchmarking robustness of earth observation foundation models}.
\newblock \bibinfo{journal}{Advances in Neural Information Processing Systems} \bibinfo{volume}{38}.
\bibitem[{Li et~al.(2024b)Li, Wen, Hu, Yuan and Zhu}]{li2024vision}
\bibinfo{author}{Li, X.}, \bibinfo{author}{Wen, C.}, \bibinfo{author}{Hu, Y.}, \bibinfo{author}{Yuan, Z.}, \bibinfo{author}{Zhu, X.X.}, \bibinfo{year}{2024}b.
\newblock \bibinfo{title}{Vision-language models in remote sensing: Current progress and future trends}.
\newblock \bibinfo{journal}{IEEE Geoscience and Remote Sensing Magazine} \bibinfo{volume}{12}, \bibinfo{pages}{32--66}.
\bibitem[{Liu et~al.(2024)Liu, Chen, Guan, Zhou, Zhu, Ye, Fu and Zhou}]{liu2024remoteclip}
\bibinfo{author}{Liu, F.}, \bibinfo{author}{Chen, D.}, \bibinfo{author}{Guan, Z.}, \bibinfo{author}{Zhou, X.}, \bibinfo{author}{Zhu, J.}, \bibinfo{author}{Ye, Q.}, \bibinfo{author}{Fu, L.}, \bibinfo{author}{Zhou, J.}, \bibinfo{year}{2024}.
\newblock \bibinfo{title}{Remoteclip: A vision language foundation model for remote sensing}.
\newblock \bibinfo{journal}{IEEE Transactions on Geoscience and Remote Sensing} \bibinfo{volume}{62}, \bibinfo{pages}{1--16}.
\bibitem[{Liu et~al.(2025)Liu, Xu, Su, Zhang and Li}]{liu2025pointsam}
\bibinfo{author}{Liu, N.}, \bibinfo{author}{Xu, X.}, \bibinfo{author}{Su, Y.}, \bibinfo{author}{Zhang, H.}, \bibinfo{author}{Li, H.C.}, \bibinfo{year}{2025}.
\newblock \bibinfo{title}{Pointsam: Pointly-supervised segment anything model for remote sensing images}.
\newblock \bibinfo{journal}{IEEE Transactions on Geoscience and Remote Sensing} \bibinfo{volume}{63}, \bibinfo{pages}{1--15}.
\bibitem[{Lu et~al.(2024)Lu, Li and Qin}]{lu2024pv}
\bibinfo{author}{Lu, N.}, \bibinfo{author}{Li, L.}, \bibinfo{author}{Qin, J.}, \bibinfo{year}{2024}.
\newblock \bibinfo{title}{Pv identifier: Extraction of small-scale distributed photovoltaics in complex environments from high spatial resolution remote sensing images}.
\newblock \bibinfo{journal}{Applied Energy} \bibinfo{volume}{365}, \bibinfo{pages}{123311}.
\bibitem[{Mahn et~al.(2024)Mahn, Kammen and Hirth}]{Mahn2024}
\bibinfo{author}{Mahn, D.}, \bibinfo{author}{Kammen, D.M.}, \bibinfo{author}{Hirth, L.}, \bibinfo{year}{2024}.
\newblock \bibinfo{title}{What drives solar energy adoption in developing countries? evidence from household-level data}.
\newblock \bibinfo{journal}{Energy Economics} \bibinfo{volume}{138}, \bibinfo{pages}{107924}.
\newblock \DOIprefix\doi{10.1016/j.eneco.2024.107924}.
\bibitem[{Malof et~al.(2016)Malof, Bradbury, Collins and Newell}]{malof2016}
\bibinfo{author}{Malof, J.M.}, \bibinfo{author}{Bradbury, K.}, \bibinfo{author}{Collins, L.M.}, \bibinfo{author}{Newell, R.G.}, \bibinfo{year}{2016}.
\newblock \bibinfo{title}{Automatic detection of solar photovoltaic arrays in high resolution aerial imagery}.
\newblock \bibinfo{journal}{Applied energy} \bibinfo{volume}{183}, \bibinfo{pages}{229--240}.
\bibitem[{Osco et~al.(2023)Osco, Wu, De~Lemos, Gon{\c{c}}alves, Ramos, Li and Junior}]{osco2023segment}
\bibinfo{author}{Osco, L.P.}, \bibinfo{author}{Wu, Q.}, \bibinfo{author}{De~Lemos, E.L.}, \bibinfo{author}{Gon{\c{c}}alves, W.N.}, \bibinfo{author}{Ramos, A.P.M.}, \bibinfo{author}{Li, J.}, \bibinfo{author}{Junior, J.M.}, \bibinfo{year}{2023}.
\newblock \bibinfo{title}{The segment anything model (sam) for remote sensing applications: From zero to one shot}.
\newblock \bibinfo{journal}{International Journal of Applied Earth Observation and Geoinformation} \bibinfo{volume}{124}, \bibinfo{pages}{103540}.
\bibitem[{Piater et~al.(2025)Piater, Barz and Freytag}]{piater2025prompt}
\bibinfo{author}{Piater, T.}, \bibinfo{author}{Barz, B.}, \bibinfo{author}{Freytag, A.}, \bibinfo{year}{2025}.
\newblock \bibinfo{title}{Prompt-tuning sam: From generalist to specialist with only 2048 parameters and 16 training images}, in: \bibinfo{booktitle}{Proceedings of the Computer Vision and Pattern Recognition Conference}, pp. \bibinfo{pages}{4688--4698}.
\bibitem[{Popp and Kalwij(2023)}]{popp2023}
\bibinfo{author}{Popp, M.R.}, \bibinfo{author}{Kalwij, J.M.}, \bibinfo{year}{2023}.
\newblock \bibinfo{title}{Consumer-grade uav imagery facilitates semantic segmentation of species-rich savanna tree layers}.
\newblock \bibinfo{journal}{Scientific Reports} \bibinfo{volume}{13}, \bibinfo{pages}{13892}.
\bibitem[{Radford et~al.(2021)Radford, Kim, Hallacy, Ramesh, Goh, Agarwal, Sastry, Askell, Mishkin, Clark et~al.}]{radford2021learning}
\bibinfo{author}{Radford, A.}, \bibinfo{author}{Kim, J.W.}, \bibinfo{author}{Hallacy, C.}, \bibinfo{author}{Ramesh, A.}, \bibinfo{author}{Goh, G.}, \bibinfo{author}{Agarwal, S.}, \bibinfo{author}{Sastry, G.}, \bibinfo{author}{Askell, A.}, \bibinfo{author}{Mishkin, P.}, \bibinfo{author}{Clark, J.}, et~al., \bibinfo{year}{2021}.
\newblock \bibinfo{title}{Learning transferable visual models from natural language supervision}, in: \bibinfo{booktitle}{International conference on machine learning}, \bibinfo{organization}{PmLR}. pp. \bibinfo{pages}{8748--8763}.
\bibitem[{Ren et~al.(2024)Ren, Luzi, Lahrichi, Kassaw, Collins, Bradbury and Malof}]{ren2024segment}
\bibinfo{author}{Ren, S.}, \bibinfo{author}{Luzi, F.}, \bibinfo{author}{Lahrichi, S.}, \bibinfo{author}{Kassaw, K.}, \bibinfo{author}{Collins, L.M.}, \bibinfo{author}{Bradbury, K.}, \bibinfo{author}{Malof, J.M.}, \bibinfo{year}{2024}.
\newblock \bibinfo{title}{Segment anything, from space?}, in: \bibinfo{booktitle}{Proceedings of the IEEE/CVF Winter Conference on Applications of Computer Vision}, pp. \bibinfo{pages}{8355--8365}.
\bibitem[{Sultan et~al.(2023)Sultan, Li, Zhu, Khanduri, Brocanelli and Zhu}]{sultan2023geosam}
\bibinfo{author}{Sultan, R.I.}, \bibinfo{author}{Li, C.}, \bibinfo{author}{Zhu, H.}, \bibinfo{author}{Khanduri, P.}, \bibinfo{author}{Brocanelli, M.}, \bibinfo{author}{Zhu, D.}, \bibinfo{year}{2023}.
\newblock \bibinfo{title}{Geosam: Fine-tuning sam with multi-modal prompts for mobility infrastructure segmentation}.
\newblock \bibinfo{journal}{arXiv preprint:2311.11319} .
\bibitem[{Tan et~al.(2024)Tan, Guo, Lin, Chen, Huang, Yuan, Zhang and Yan}]{tan2024general}
\bibinfo{author}{Tan, H.}, \bibinfo{author}{Guo, Z.}, \bibinfo{author}{Lin, Z.}, \bibinfo{author}{Chen, Y.}, \bibinfo{author}{Huang, D.}, \bibinfo{author}{Yuan, W.}, \bibinfo{author}{Zhang, H.}, \bibinfo{author}{Yan, J.}, \bibinfo{year}{2024}.
\newblock \bibinfo{title}{General generative ai-based image augmentation method for robust rooftop pv segmentation}.
\newblock \bibinfo{journal}{Applied Energy} \bibinfo{volume}{368}, \bibinfo{pages}{123554}.
\bibitem[{Tan et~al.(2023)Tan, Guo, Zhang, Chen, Lin, Chen and Yan}]{tan2023enhancing}
\bibinfo{author}{Tan, H.}, \bibinfo{author}{Guo, Z.}, \bibinfo{author}{Zhang, H.}, \bibinfo{author}{Chen, Q.}, \bibinfo{author}{Lin, Z.}, \bibinfo{author}{Chen, Y.}, \bibinfo{author}{Yan, J.}, \bibinfo{year}{2023}.
\newblock \bibinfo{title}{Enhancing pv panel segmentation in remote sensing images with constraint refinement modules}.
\newblock \bibinfo{journal}{Applied Energy} \bibinfo{volume}{350}, \bibinfo{pages}{121757}.
\bibitem[{Vrande{\v{c}}i{\'c} and Kr{\"o}tzsch(2014)}]{vrandevcic2014}
\bibinfo{author}{Vrande{\v{c}}i{\'c}, D.}, \bibinfo{author}{Kr{\"o}tzsch, M.}, \bibinfo{year}{2014}.
\newblock \bibinfo{title}{Wikidata: a free collaborative knowledgebase}.
\newblock \bibinfo{journal}{Communications of the ACM} \bibinfo{volume}{57}, \bibinfo{pages}{78--85}.
\bibitem[{Wang et~al.(2023)Wang, Zhang, Du, Xu, Liu, Tao and Zhang}]{wang2023samrs}
\bibinfo{author}{Wang, D.}, \bibinfo{author}{Zhang, J.}, \bibinfo{author}{Du, B.}, \bibinfo{author}{Xu, M.}, \bibinfo{author}{Liu, L.}, \bibinfo{author}{Tao, D.}, \bibinfo{author}{Zhang, L.}, \bibinfo{year}{2023}.
\newblock \bibinfo{title}{Samrs: Scaling-up remote sensing segmentation dataset with segment anything model}.
\newblock \bibinfo{journal}{Advances in Neural Information Processing Systems} \bibinfo{volume}{36}, \bibinfo{pages}{8815--8827}.
\bibitem[{Wang et~al.(2025)Wang, Shao, Hou and Cai}]{wang2025pv}
\bibinfo{author}{Wang, S.}, \bibinfo{author}{Shao, Z.}, \bibinfo{author}{Hou, D.}, \bibinfo{author}{Cai, B.}, \bibinfo{year}{2025}.
\newblock \bibinfo{title}{Pv segmenter: A frequency-guided edge-aware network for distributed photovoltaic segmentation in remote sensing imagery}.
\newblock \bibinfo{journal}{Applied Energy} \bibinfo{volume}{393}, \bibinfo{pages}{126137}.
\bibitem[{Wang et~al.(2022)Wang, Lu, Li, Tao, Guo, Gong and Liu}]{wang2022cris}
\bibinfo{author}{Wang, Z.}, \bibinfo{author}{Lu, Y.}, \bibinfo{author}{Li, Q.}, \bibinfo{author}{Tao, X.}, \bibinfo{author}{Guo, Y.}, \bibinfo{author}{Gong, M.}, \bibinfo{author}{Liu, T.}, \bibinfo{year}{2022}.
\newblock \bibinfo{title}{Cris: Clip-driven referring image segmentation}, in: \bibinfo{booktitle}{Proceedings of the IEEE/CVF conference on computer vision and pattern recognition}, pp. \bibinfo{pages}{11686--11695}.
\bibitem[{Wortsman et~al.(2022)Wortsman, Ilharco, Kim, Li, Kornblith, Roelofs, Lopes, Hajishirzi, Farhadi, Namkoong et~al.}]{wortsman2022}
\bibinfo{author}{Wortsman, M.}, \bibinfo{author}{Ilharco, G.}, \bibinfo{author}{Kim, J.W.}, \bibinfo{author}{Li, M.}, \bibinfo{author}{Kornblith, S.}, \bibinfo{author}{Roelofs, R.}, \bibinfo{author}{Lopes, R.G.}, \bibinfo{author}{Hajishirzi, H.}, \bibinfo{author}{Farhadi, A.}, \bibinfo{author}{Namkoong, H.}, et~al., \bibinfo{year}{2022}.
\newblock \bibinfo{title}{Robust fine-tuning of zero-shot models}, in: \bibinfo{booktitle}{Proceedings of the IEEE/CVF conference on computer vision and pattern recognition}, pp. \bibinfo{pages}{7959--7971}.
\bibitem[{Xie et~al.(2021)Xie, Wang, Yu, Anandkumar, Alvarez and Luo}]{xie2021segformer}
\bibinfo{author}{Xie, E.}, \bibinfo{author}{Wang, W.}, \bibinfo{author}{Yu, Z.}, \bibinfo{author}{Anandkumar, A.}, \bibinfo{author}{Alvarez, J.M.}, \bibinfo{author}{Luo, P.}, \bibinfo{year}{2021}.
\newblock \bibinfo{title}{Segformer: Simple and efficient design for semantic segmentation with transformers}.
\newblock \bibinfo{journal}{Advances in neural information processing systems} \bibinfo{volume}{34}, \bibinfo{pages}{12077--12090}.
\bibitem[{Xin et~al.(2025)Xin, Li, Chen, Li, Xiao, Qiao, Zhang, Meng and Cao}]{xin2025segearth}
\bibinfo{author}{Xin, Z.}, \bibinfo{author}{Li, K.}, \bibinfo{author}{Chen, L.}, \bibinfo{author}{Li, W.}, \bibinfo{author}{Xiao, Y.}, \bibinfo{author}{Qiao, H.}, \bibinfo{author}{Zhang, W.}, \bibinfo{author}{Meng, D.}, \bibinfo{author}{Cao, X.}, \bibinfo{year}{2025}.
\newblock \bibinfo{title}{Segearth-r2:towards comprehensive language-guided segmentation for remote sensing images}.
\newblock \bibinfo{journal}{arXiv preprint:2512.20013} .
\bibitem[{Yao et~al.(2025)Yao, Liu, Chen, Zhang, Wang, Chen, Xu, Di and Zheng}]{yao2025remotesam}
\bibinfo{author}{Yao, L.}, \bibinfo{author}{Liu, F.}, \bibinfo{author}{Chen, D.}, \bibinfo{author}{Zhang, C.}, \bibinfo{author}{Wang, Y.}, \bibinfo{author}{Chen, Z.}, \bibinfo{author}{Xu, W.}, \bibinfo{author}{Di, S.}, \bibinfo{author}{Zheng, Y.}, \bibinfo{year}{2025}.
\newblock \bibinfo{title}{Remotesam: Towards segment anything for earth observation}, in: \bibinfo{booktitle}{Proceedings of the 33rd ACM International Conference on Multimedia}, pp. \bibinfo{pages}{3027--3036}.
\bibitem[{Yao et~al.(2026)Yao, Zhang, Liang, Li and Liu}]{yao2026pvsam}
\bibinfo{author}{Yao, X.}, \bibinfo{author}{Zhang, S.}, \bibinfo{author}{Liang, Z.}, \bibinfo{author}{Li, J.}, \bibinfo{author}{Liu, C.}, \bibinfo{year}{2026}.
\newblock \bibinfo{title}{Pvsam: Adapting geometric prompts to segment anything model for photovoltaic detection in remote sensing imagery}.
\newblock \bibinfo{journal}{Applied Energy} \bibinfo{volume}{404}, \bibinfo{pages}{127137}.
\bibitem[{You et~al.(2024)You, Mint, Dai, Sekhon, Staib and Duncan}]{you2024}
\bibinfo{author}{You, C.}, \bibinfo{author}{Mint, Y.}, \bibinfo{author}{Dai, W.}, \bibinfo{author}{Sekhon, J.S.}, \bibinfo{author}{Staib, L.}, \bibinfo{author}{Duncan, J.S.}, \bibinfo{year}{2024}.
\newblock \bibinfo{title}{Calibrating multi-modal representations: A pursuit of group robustness without annotations}, in: \bibinfo{booktitle}{2024 IEEE/CVF Conference on Computer Vision and Pattern Recognition (CVPR)}, \bibinfo{organization}{IEEE}. pp. \bibinfo{pages}{26140--26150}.
\bibitem[{Yu et~al.(2018)Yu, Wang, Majumdar and Rajagopal}]{yu2018deepsolar}
\bibinfo{author}{Yu, J.}, \bibinfo{author}{Wang, Z.}, \bibinfo{author}{Majumdar, A.}, \bibinfo{author}{Rajagopal, R.}, \bibinfo{year}{2018}.
\newblock \bibinfo{title}{Deepsolar: A machine learning framework to efficiently construct a solar deployment database in the united states}.
\newblock \bibinfo{journal}{Joule} \bibinfo{volume}{2}, \bibinfo{pages}{2605--2617}.
\bibitem[{Yuan et~al.(2025)Yuan, Xiong, Mou and Zhu}]{yuan2025}
\bibinfo{author}{Yuan, Z.}, \bibinfo{author}{Xiong, Z.}, \bibinfo{author}{Mou, L.}, \bibinfo{author}{Zhu, X.X.}, \bibinfo{year}{2025}.
\newblock \bibinfo{title}{Chatearthnet: A global-scale image-text dataset empowering vision-language geo-foundation models}.
\newblock \bibinfo{journal}{Earth System Science Data Discussions} \bibinfo{volume}{2024}, \bibinfo{pages}{1--24}.
\bibitem[{Zech et~al.(2024)Zech, Tetens and Ranalli}]{zech2024toward}
\bibinfo{author}{Zech, M.}, \bibinfo{author}{Tetens, H.P.}, \bibinfo{author}{Ranalli, J.}, \bibinfo{year}{2024}.
\newblock \bibinfo{title}{Toward global rooftop pv detection with deep active learning}.
\newblock \bibinfo{journal}{Advances in Applied Energy} \bibinfo{volume}{16}, \bibinfo{pages}{100191}.
\bibitem[{Zhang et~al.(2025)Zhang, Li, Yang, Jiang and Zhang}]{zhang2025rsam}
\bibinfo{author}{Zhang, J.}, \bibinfo{author}{Li, Y.}, \bibinfo{author}{Yang, X.}, \bibinfo{author}{Jiang, R.}, \bibinfo{author}{Zhang, L.}, \bibinfo{year}{2025}.
\newblock \bibinfo{title}{Rsam-seg: A sam-based model with prior knowledge integration for remote sensing image semantic segmentation}.
\newblock \bibinfo{journal}{Remote Sensing} \bibinfo{volume}{17}, \bibinfo{pages}{590}.
\bibitem[{Zhang et~al.(2024a)Zhang, Zhou, Mai, Hu, Guan, Li and Mu}]{zhang2024T2S}
\bibinfo{author}{Zhang, J.}, \bibinfo{author}{Zhou, Z.}, \bibinfo{author}{Mai, G.}, \bibinfo{author}{Hu, M.}, \bibinfo{author}{Guan, Z.}, \bibinfo{author}{Li, S.}, \bibinfo{author}{Mu, L.}, \bibinfo{year}{2024}a.
\newblock \bibinfo{title}{Text2seg: Zero-shot remote sensing image semantic segmentation via text-guided visual foundation models}, in: \bibinfo{booktitle}{Proceedings of the 7th ACM SIGSPATIAL International workshop on AI for geographic knowledge discovery}, pp. \bibinfo{pages}{63--66}.
\bibitem[{Zhang et~al.(2024b)Zhang, Zhang, Wu, Zhou, Jiang and Ma}]{zhang2024segclip}
\bibinfo{author}{Zhang, S.}, \bibinfo{author}{Zhang, B.}, \bibinfo{author}{Wu, Y.}, \bibinfo{author}{Zhou, H.}, \bibinfo{author}{Jiang, J.}, \bibinfo{author}{Ma, J.}, \bibinfo{year}{2024}b.
\newblock \bibinfo{title}{Segclip: Multimodal visual-language and prompt learning for high-resolution remote sensing semantic segmentation}.
\newblock \bibinfo{journal}{IEEE Transactions on Geoscience and Remote Sensing} .
\bibitem[{Zhang et~al.(2024c)Zhang, Zhao, Guo and Yin}]{zhang2024rs5m}
\bibinfo{author}{Zhang, Z.}, \bibinfo{author}{Zhao, T.}, \bibinfo{author}{Guo, Y.}, \bibinfo{author}{Yin, J.}, \bibinfo{year}{2024}c.
\newblock \bibinfo{title}{Rs5m and georsclip: A large-scale vision-language dataset and a large vision-language model for remote sensing}.
\newblock \bibinfo{journal}{IEEE Transactions on Geoscience and Remote Sensing} \bibinfo{volume}{62}, \bibinfo{pages}{1--23}.
\bibitem[{Zhao et~al.(2024)Zhao, Alzubaidi, Zhang, Duan and Gu}]{zhao2024comparison}
\bibinfo{author}{Zhao, Z.}, \bibinfo{author}{Alzubaidi, L.}, \bibinfo{author}{Zhang, J.}, \bibinfo{author}{Duan, Y.}, \bibinfo{author}{Gu, Y.}, \bibinfo{year}{2024}.
\newblock \bibinfo{title}{A comparison review of transfer learning and self-supervised learning: Definitions, applications, advantages and limitations}.
\newblock \bibinfo{journal}{Expert Systems with Applications} \bibinfo{volume}{242}, \bibinfo{pages}{122807}.
\bibitem[{Zhao et~al.(2025)Zhao, Li, Chen and Wang}]{zhao2025enhancing}
\bibinfo{author}{Zhao, Z.}, \bibinfo{author}{Li, K.}, \bibinfo{author}{Chen, Y.}, \bibinfo{author}{Wang, J.}, \bibinfo{year}{2025}.
\newblock \bibinfo{title}{Enhancing visual feature constraints in segmentation models for photovoltaic panel recognition}.
\newblock \bibinfo{journal}{Energy and AI} \bibinfo{volume}{21}, \bibinfo{pages}{100544}.
\bibitem[{Zhou et~al.(2024)Zhou, Lan, Li, Feng, Ke, Jiang, Li, Yang and Zhang}]{zhou2024geo}
\bibinfo{author}{Zhou, Y.}, \bibinfo{author}{Lan, M.}, \bibinfo{author}{Li, X.}, \bibinfo{author}{Feng, L.}, \bibinfo{author}{Ke, Y.}, \bibinfo{author}{Jiang, X.}, \bibinfo{author}{Li, Q.}, \bibinfo{author}{Yang, X.}, \bibinfo{author}{Zhang, W.}, \bibinfo{year}{2024}.
\newblock \bibinfo{title}{Geoground: A unified large vision-language model for remote sensing visual grounding}.
\newblock \bibinfo{journal}{arXiv preprint:2411.11904} .
\bibitem[{Zhu et~al.(2017)Zhu, Tuia, Mou, Xia, Zhang, Xu and Fraundorfer}]{zhu2017deep}
\bibinfo{author}{Zhu, X.X.}, \bibinfo{author}{Tuia, D.}, \bibinfo{author}{Mou, L.}, \bibinfo{author}{Xia, G.S.}, \bibinfo{author}{Zhang, L.}, \bibinfo{author}{Xu, F.}, \bibinfo{author}{Fraundorfer, F.}, \bibinfo{year}{2017}.
\newblock \bibinfo{title}{Deep learning in remote sensing: A comprehensive review and list of resources}.
\newblock \bibinfo{journal}{IEEE geoscience and remote sensing magazine} \bibinfo{volume}{5}, \bibinfo{pages}{8--36}.

\end{thebibliography}

\renewcommand{\thetable}{S\arabic{table}}
\setcounter{table}{0}
\section*{Supplementary Tables}

\begin{table}[H]
\centering
\caption{\textbf{Type-II ANOVA results for year effects on F1 score.}
Year effects were estimated separately for each prompting strategy and spatial resolution, while controlling for training size and training strategy.}
\label{anova}
\footnotesize
\renewcommand{\arraystretch}{1.15}
\setlength{\tabcolsep}{5.5pt}
\begin{tabular}{llcccccc}
\toprule
\textbf{Resolution} & \textbf{Prompt} & \textbf{N} & \textbf{df$_{\mathrm{y}}$} & \textbf{df$_{\mathrm{r}}$} & \textbf{F} & \textbf{p-value} & \textbf{$\eta_p^2$} \\
\midrule
0.250 & Textual      & 42 & 2 & 32 & 33.84 & $<0.001$ & 0.679 \\
0.250 & Geometric & 42 & 2 & 32 & 8.85  & $<0.001$ & 0.356 \\
0.250 & Hybrid    & 42 & 2 & 32 & 4.97  & 0.013    & 0.237 \\
\midrule
0.145 & Textual      & 28 & 1 & 19 & 1.93  & 0.181    & 0.092 \\
0.145 & Geometric & 28 & 1 & 19 & 1.23  & 0.280    & 0.061 \\
0.145 & Hybrid    & 28 & 1 & 19 & 0.19  & 0.666    & 0.010 \\
\midrule
0.132 & Textual      & 28 & 1 & 19 & 2.42  & 0.136    & 0.113 \\
0.132 & Geometric & 28 & 1 & 19 & 2.06  & 0.167    & 0.098 \\
0.132 & Hybrid    & 28 & 1 & 19 & 1.68  & 0.211    & 0.081 \\
\bottomrule
\end{tabular}
\begin{center} 
\vspace{-3pt}
\footnotesize $\eta_p^2$ = partial eta squared; N = years $\times$ training regimes $\times$ supervision scales; {df$_{\mathrm{y}}$} = years degrees of freedom. df$_{\mathrm{r}}$ = residual degrees of freedom.
\end{center}
\end{table}
\vspace{-6pt}
\begin{table}[H]
\centering
\caption{\textbf{Radiometric characteristics of the aerial imagery, by year.}
Intensity was computed as the mean of the RGB channels. Dynamic range was calculated as $DR=P_{95}-P_{5}$.}
\label{radiometry}
\footnotesize
\renewcommand{\arraystretch}{1.15}
\setlength{\tabcolsep}{6pt}

\begin{tabular}{ccccccc}
\toprule
\textbf{Year} & \textbf{Resolution} & \textbf{Mean} & \textbf{SD} & \textbf{$P_5$} & \textbf{$P_{95}$} & \textbf{DR} \\
\midrule
2012 & 0.250 & 95.3  & 25.5 & 47.6 & 123.0 & 75.4  \\
2015 & 0.250 & 109.9 & 25.4 & 59.3 & 143.3 & 84.0  \\
2016 & 0.250 & 126.2 & 50.1 & 36.0 & 197.6 & 161.6 \\
2017 & 0.145 & 144.0 & 34.8 & 76.6 & 209.8 & 133.3 \\
2020 & 0.145 & 116.0 & 37.1 & 62.5 & 196.2 & 131.6 \\
2021 & 0.132 & 121.3 & 33.6 & 55.5 & 185.3 & 129.8 \\
2022 & 0.132 & 130.4 & 40.8 & 50.7 & 208.8 & 158.1 \\
\bottomrule
\end{tabular}
\end{table}

\begin{table}[H]
\centering
\caption{\textbf{Performance improvements between comparable-resolution datasets}. Mean performance differences ($\Delta$) between the Queens benchmark dataset (0.15 m/pixel) and the primary dataset (0.145 m/pixel; 2017 and 2020 imagery). Values were averaged across all supervision scales using independently trained models only (ZS excluded).}
\label{queens}
\renewcommand{\arraystretch}{1.15}
\setlength{\tabcolsep}{8pt}
\begin{tabular}{lccc}
\toprule
\textbf{Prompting} & $\Delta$Precision & $\Delta$Recall & $\Delta$IoU \\
\midrule
Textual   & +0.076 & +0.146 & +0.167 \\
Geometric & +0.064 & +0.152 & +0.176 \\
Hybrid    & +0.029 & +0.035 & +0.057 \\
\bottomrule
\end{tabular}
\end{table}

\end{document}